\pdfoutput=1

\documentclass[11pt]{article}

\usepackage[preprint]{acl}

\usepackage{times}
\usepackage{latexsym}

\usepackage[T1]{fontenc}

\usepackage[utf8]{inputenc}

\usepackage{microtype}

\usepackage{inconsolata}

\usepackage{graphicx}
\usepackage{booktabs}
\usepackage{multirow}
\usepackage{enumitem}

\usepackage{hyperref}
\usepackage{fontawesome5}

\usepackage{tabularx}
\usepackage[dvipsnames]{xcolor} 
\usepackage[most]{tcolorbox}    
\usepackage{amssymb}            
\usepackage{pifont}             
\usepackage{longtable}
\usepackage{xspace}
\usepackage[normalem]{ulem}
\usepackage{amssymb}
\usepackage{titletoc}
\usepackage{subcaption}

\newcommand{\ie}{\hbox{\emph{i.e.,}}\xspace}

\newcommand{\ours}{MedTutor\xspace}

\usepackage{lipsum} 

\newcommand{\hfmark}{\raisebox{-0.7ex}{\includegraphics[height=1.35em]{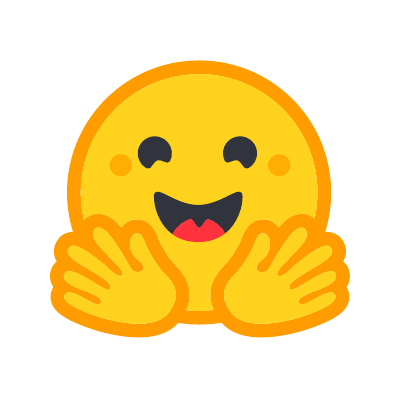}}}

\definecolor{youtubeRed}{HTML}{FF0000}
\newcommand{\yticon}{\textcolor{youtubeRed}{\faYoutube}}

\title{\ours: A Retrieval-Augmented LLM System for\\ Case-Based Medical Education}

\author{
  \textbf{Dongsuk Jang\textsuperscript{1,3}}\quad
  \textbf{Ziyao Shangguan\textsuperscript{1}}\quad
  \textbf{Kyle Tegtmeyer\textsuperscript{2}}\quad\\
  \textbf{Anurag Gupta\textsuperscript{2}}\quad
  \textbf{Jan Czerminski\textsuperscript{2}}\quad
    \textbf{Sophie Chheang\textsuperscript{2}}\quad
  \textbf{Arman Cohan\textsuperscript{1}}\\
  \textsuperscript{1}Department of Computer Science, Yale University \\
  \textsuperscript{2}Department of Radiology and Biomedical Imaging, Yale School of Medicine \\
  \textsuperscript{3}Interdisciplinary Program for Bioengineering, Seoul National University \\
  \texttt{\{james.jang,ziyao.shangguan,arman.cohan\}@yale.edu} \\ 
      \faGithub~\href{https://github.com/yale-nlp/medical-rag}{Code}\hspace{1.5em}
      \yticon~\href{https://www.youtube.com/watch?v=7NlCjVf8V4E}{Demo Video}\hspace{1.5em}
      \href{https://huggingface.co/datasets/yale-nlp/MedTutor}{\hfmark~Dataset}
}

\begin{document}
\maketitle
\begin{abstract}
The learning process for medical residents presents significant challenges, demanding both the ability to interpret complex case reports and the rapid acquisition of accurate medical knowledge from reliable sources.
Residents typically study case reports and engage in discussions with peers and mentors, but finding relevant educational materials and evidence to support their learning from these cases is often time-consuming and challenging.
To address this, we introduce \textbf{\ours}, a novel system designed to augment resident training by automatically generating evidence-based educational content and multiple-choice questions from clinical case reports. \ours leverages a Retrieval-Augmented Generation (RAG) pipeline that takes clinical case reports as input and produces targeted educational materials. The system's architecture features a hybrid retrieval mechanism that synergistically queries a local knowledge base of medical textbooks and academic literature (using PubMed, Semantic Scholar APIs) for the latest related research, ensuring the generated content is both foundationally sound and current. The retrieved evidence is filtered and ordered using a state-of-the-art reranking model and then an LLM generates the final long-form output describing the main educational content regarding the case-report.
We conduct a rigorous evaluation of the system. First, three radiologists assessed the quality of outputs, finding them to be of high clinical and educational value. Second, we perform a large-scale evaluation using an LLM-as-a Judge to understand if LLMs can be used to evaluate the output of the system. Our analysis using correlation between LLMs outputs and human expert judgments reveals a moderate alignment and highlights the continued necessity of expert oversight. 

\end{abstract}

\section{Introduction}

The training of medical residents is an intensive learning process, built upon the foundation of studying and interpreting thousands of case reports. Residents routinely engage with clinical cases through discussions with peers and mentors, analyzing findings and differential diagnoses to deepen their understanding. However, while direct feedback from attending physicians is invaluable, the process of finding relevant educational material and supporting evidence for specific cases is often time-consuming and inconsistent \citep{Rogers2019EducationalRU, Daniel2020PediatricRE}. The sheer volume of medical literature and the challenge of identifying pertinent resources for each case can limit the depth of learning that residents achieve from their clinical experiences \citep{Anderson2019HowSA, Bednarczyk2014CharacteristicsOE}. There exists a significant opportunity to augment this traditional learning process with AI tools that can efficiently retrieve and synthesize educational content from clinical cases, drawing upon vast archives of medical knowledge. LLMs present a promising avenue for this augmentation, but their application in high-stakes medical domains is fraught with challenges, most notably the risk of factual inaccuracy (or hallucination) and the use of outdated knowledge \citep{Abd-alrazaq2023LargeLM, Li2023ASO, Xie2023FaithfulAI}.

\begin{figure*}[t]
    \centering
    \includegraphics[width=\textwidth]{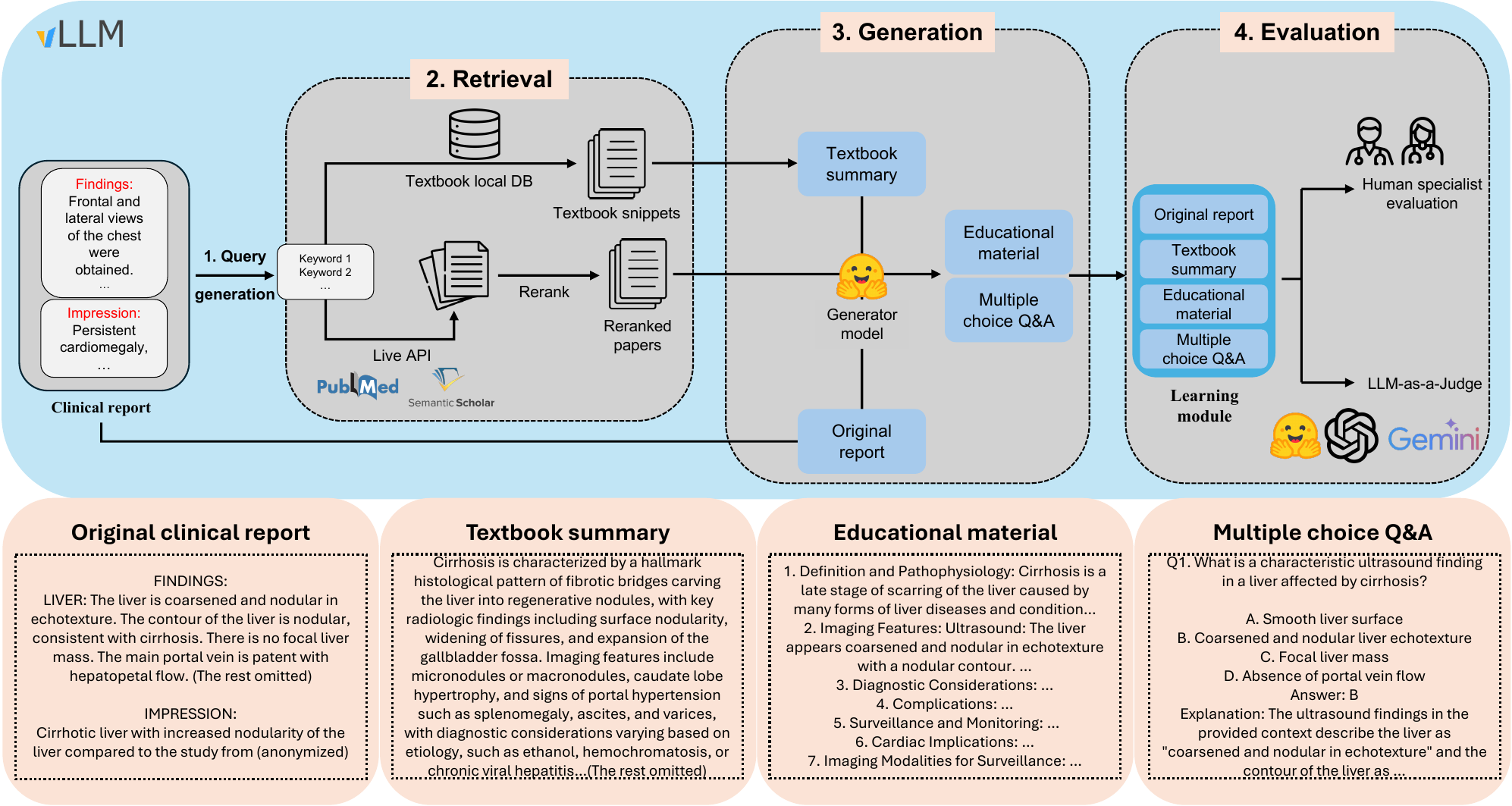}
    \caption{The overall architecture of the \ours system.}
    \label{fig:architecture}
\end{figure*}

To overcome these challenges, we develop \ours, a system that grounds LLM generation in verifiable, contextually relevant medical knowledge to case reports through a RAG pipeline. Our primary goal is to provide medical residents with a reliable tool that transforms any given clinical report into a concise, and highly relevant educational module. We focus on radiology as the domain of study, although, our techniques are generalizable to other domains. The system begins by decomposing clinical reports into actionable diagnostic queries and keywords that can be effectively issued to a search index, enabling targeted retrieval of relevant educational material. It then initiates a hybrid retrieval process that simultaneously queries a curated database of medical textbooks, and performs live searches on academic search engines (i.e., PubMed and Semantic Scholar) for current published literature related to the case.

The retrieved evidence undergoes a multi-faceted processing step: academic articles and textbook snippets are reranked for relevance to the case, using a state-of-the-art reranking model, Qwen3-Reranker-8B \citep{zhang2025qwen3embeddingadvancingtext}. Finally, all processed evidence---the original report, keywords, top-ranked articles, and textbook summaries---is synthesized by a generator LLM into two distinct outputs: a comprehensive set of educational material and a set of multiple-choice questions (MCQs) designed to test understanding. The overview of the system is illustrated in Figure \ref{fig:architecture}.

This work makes three primary contributions: \vspace{-6pt}
\begin{itemize}[wide,leftmargin=0pt,noitemsep]
    \item We detail the design and implementation of the \ours system, a scalable and efficient architecture that leverages asynchronous I/O, parallel multi-GPU inference with vLLM \citep{kwon2023efficient}, and optimized batch processing to handle large workloads. 
    \item We introduce a new, expert annotated benchmark dataset for evaluating the quality of AI-generated educational content. We run our pipeline with 6 LLMs(see Appendix~\ref{sec:llm_as_judge_eval} for details) across 2,000 clinical reports per each 5 major radiology datasets (\ie Yale Hospital Internal, MIMIC-CXR \citep{johnson2019mimic, johnson2024mimiccxr}, MIMIC-IV-note \citep{johnson2023mimic}, CheXpert Plus \citep{chambon2024chexpertplusaugmentinglarge}, and ReXGradient-160K \citep{zhang2025rexgradient160klargescalepubliclyavailable}).
    \item We collected comprehensive evaluations from three radiologists, alongside a LLM-as-a-Judge evaluation with four models for all experiments. This dataset, which we are planning to publicly release, will be a valuable resource for evaluating the quality and clinical utility of generative models in medicine. 
\end{itemize}

Our analysis provides insights about the usefulness of our system to users and highlight the strengths and weaknesses of LLMs in evaluating educational content in our setting.

\section{Related Work}
\label{sec:related_works}

Our work is situated at the intersection of RAG, the application of LLMs in medicine, and the critical need for trustworthy medical AI systems. We structure our review accordingly.

\subsection{LLMs and RAG in the Medical Domain}
The application of LLMs in medicine has shown immense promise. General-purpose foundation models have demonstrated impressive capabilities on standardized medical exams and complex diagnostic problems \citep{nori2023capabilitiesgpt4medicalchallenge, singhal2023expertlevelmedicalquestionanswering}. This has fueled a broader vision for generalist biomedical AI that can assist with a wide range of clinical tasks \citep{tu2023generalistbiomedicalai}. However, the ``black-box'' nature of these models and their potential for factual errors or "hallucinations" remain significant barriers to clinical adoption, necessitating robust evaluation frameworks \citep{huang2024survey, Li2023ASO, Xie2023FaithfulAI}.

To mitigate these risks, RAG has emerged as a key paradigm for building dependable clinical tools \citep{lewis2021retrievalaugmentedgenerationknowledgeintensivenlp}. By grounding LLM outputs in external, verifiable evidence from reliable medical literature, RAG provides a pathway to trustworthy AI. A recent perspective in \textit{Nature Medicine} strongly advocates for RAG as a prerequisite for the responsible deployment of generative AI in healthcare \citep{yang2024retrievalaugmentedgenerationgenerativeartificial}. The field is now maturing to a point where standardized benchmarks for medical RAG are being established, allowing for more rigorous evaluation of these systems \citep{xiong2024benchmarking}. 
Our work contributes to this growing body of literature by presenting a novel RAG-based system specifically designed for medical education, a domain where accuracy and reliability are paramount.

\subsection{LLMs for Medical Education}
LLMs show promise in generating medical exam content, though concerns about accuracy necessitate expert oversight \cite{zhu2024potential}. Integrating RAG improves reliability by grounding output in external sources, with studies reporting notable gains in question-answering accuracy using medical textbooks \cite{chen2025enhancing, wang2023blackbox}. Benchmarks like MIRAGE further validate RAG's role in medical QA tasks \cite{xiong2024benchmarking}. 

For resident training, LLMs can assess skills and provide feedback, but expert review remains vital \cite{atsukawa2024evaluation}. Systems enabling citation generation enhance factuality \cite{wang2025medcite}, while evaluation frameworks like LLM-as-a-Judge offer scalability despite only moderate alignment with human judgment \cite{zheng2025fuzzyjudge}. New approaches continue to embed evidence-based medicine principles into RAG pipelines for clinically accurate educational content \cite{lu2025medr2}. 

Our system, \ours, is distinct in its focus on transforming a single clinical report into a comprehensive educational module, featuring synthesized educational material and MCQs grounded in a hybrid retrieval from both medical textbooks and the latest academic literature. This approach is designed not to replace expert judgment but to augment it, fostering the self-directed learning skills that are crucial for lifelong professional development \citep{Bravata2003TheDA, Williams2018AnOS}.

\section{\ours}
\label{sec:system_design}

\ours is a RAG system designed to support medical residents on case-based education. It involves a pipeline approach in retrieving highly relevant educational content from both textbooks and literature and produces a coherent educational material as well as multiple-choice questions related to a case. While \ours's design is general and can be applicable to many clinical practices, we focus our domain on radiology due to availability of public datasets and our access to domain experts.  

\subsection{The \ours  Pipeline Stages}

The input to \ours is a case report, which will be processed through a sequence of automated stages, each designed for parallel execution.

\textbf{Case decomposition into search queries:} The process begins with a source radiology report. Then we use an Llama-3.3-70B-Instruct\citep{meta_llama3_3_70b_instruct} to process the radiology report and decompose it into multiple keyword based queries that will be used for retrieval. These queries are key diagnostic terms and findings. Prompts for case decomposition into search queries are shown in Appendix \ref{sec: keyword generation prompt}.

\textbf{Hybrid Evidence Retrieval:} For each search query, the system performs a hybrid retrieval process in parallel described below:
\emph{(1) Local retrieval for textbook material:} Textbooks and notes are essential resources for medical education. In our \ours system, we first apply OCR to a radiology textbook \citep{wolfgang2017} using the \texttt{mistral-ocr-2503} model, then segment and index the material by page. We generate dense embeddings for these materials with the \texttt{Qwen3-Embedding-8B} model, which has demonstrated state-of-the-art performance in embedding and retrieval tasks on the MTEB benchmark \citep{muennighoff2022mteb} among models of comparable size. These embeddings are stored in a pre-computed vector database for subsequent queries. For local database search, we employ a bi-encoder architecture to generate dense vector representations for both the query and the pre-indexed textbook pages, subsequently identifying the most relevant page using cosine similarity.
\emph{(2) Retrieval using academic APIs:} Some case reports are more specialized or rare, requiring retrieving knowledge from latest academic literature. Therefore, we also employ retrieval from academic search engines. We use PubMed and Semantic Scholar APIs, two commonly used and freely available scholarly systems, to fetch the latest relevant research papers. To prevent rate-limiting, API calls are managed by an \texttt{asyncio.Semaphore}. If pre-fetched results for the queries are available, this step is skipped to improve efficiency.

\textbf{Evidence Processing:} The retrieved evidence is then processed through two concurrent tasks:
\emph{(1) Reranking:} As the search engine results using keyword queries can be noisy, we employ a reranking stage to prioritize the most relevant (top 2) documents to the case report. This is handled by a dedicated service running the Qwen3-Reranker-8B model, a strong reranker according to the MTEB benchmark. The reranker is given a contextualized query containing both the original report's text and the specific search keyword to improve relevance.
\emph{(2) Query-focused Summarization:} Concurrently, the content retrieved from the local textbook database is summarized with respect to the query by a generator LLM to distill key information related to the keywords into a concise way.

\textbf{Generating Learning Modules:} Finally, the original case report, the top retrieved content including the textbook snippets and abstracts of related papers, and the search keywords are passed to a generator LLM to generate a concise learning module.  These learning modules contain comprehensive explanatory material contextualizing the case within broader medical knowledge, followed by multiple-choice questions designed to test understanding of key concepts. Prompts used for generating learning modules are in Appendix~\ref{sec:appendix_prompts}.

\textbf{Optimized Multi-Task Generation:} The generation step is heavily optimized for efficiency. Instead of generating outputs sequentially, the system first constructs prompts for all cases received, and all sub-tasks. 

\emph{Batch Construction:} Two distinct batches of input prompts to LLMs are created: one for generating the final educational modules and another for generating multiple-choice questions. These input prompts are long-context (3530 tokens for MCQ, 3463 tokens for Educational module in average), containing the original report, the list of keywords, the abstracts of the top-ranked papers after reranking, and the generated textbook summaries.
\textit{Concurrent Batch Inference:} The two batches are sent concurrently to the generation service. The \texttt{generate\_text\_batch} method in our \texttt{VLLMHandler} passes the entire list of prompts to the vLLM engine in a single call. This fully leverages vLLM's continuous batching capability, allowing the GPU to process multiple requests simultaneously without padding, dramatically increasing throughput and reducing overall processing time.
This architecture, particularly the use of batch generation with vLLM, allows \ours to process hundreds of complex reports far more efficiently than a naive, sequential approach, making it a practical tool for large-scale educational content creation.

\textbf{Local Deployment:} We deploy \ours completely locally using locally served open-source LLMs, without reliance on any cloud-based LLM APIs. This allows responsible and private handling of medical data.

\subsection{System Design Details}

The \ours pipeline is an asynchronous, multi-stage system designed for efficiency, scalability, and modularity. The architecture leverages parallel processing across multiple GPUs and optimized batching to handle large-scale report generation. The entire workflow is orchestrated by a central \texttt{asyncio} event loop, which communicates with dedicated \texttt{ModelWorker} processes via multiprocessing queues. A conceptual overview of the architecture is shown in Figure~\ref{fig:architecture}.

\subsection{Architecture for Scalability}

At the core of our system is a hybrid concurrency model designed to maximize throughput and resource utilization.

\textbf{Asynchronous Orchestration:} The main process runs on an \texttt{asyncio} event loop, managing I/O-bound tasks such as live API calls for literature retrieval and orchestrating the overall pipeline. This allows the system to handle thousands of concurrent operations efficiently without being blocked by network latency.

\textbf{Parallel Multi-GPU Inference:} To handle the computationally intensive model inference, we spawn separate \texttt{ModelWorker} processes for each required service (e.g., reranking, generation). Each worker is pinned to a specific GPU or set of GPUs as defined in the \texttt{configs.json} file. Within each worker, we use the vLLM engine, a state-of-the-art serving library that employs techniques like PagedAttention to achieve high-throughput, low-latency inference.

\textbf{Inter-Process Communication:} The main \texttt{asyncio} loop communicates with the \texttt{ModelWorker} processes using a robust queue-based system (\texttt{multiprocessing.Queue}). A request, tagged with a unique ID, is placed on a request queue. The main loop then awaits an \texttt{asyncio.Future} associated with that ID. The worker process retrieves the request, performs the inference, and places the result on a response queue. A dedicated listener task in the main loop listens for responses and resolves the corresponding \texttt{Future}, seamlessly bridging the asynchronous and multi-process components.

\section{System Evaluation}
\label{sec:system_evaluation}

We conduct a multi-faceted evaluation to assess the quality of our \ours system. Given our focus on the radiology domain, the evaluation is done by three radiologists who scored the outputs on a 5-point Likert scale (1=Poor, 5=Excellent).
Annotation guidelines and the annotation interface design are detailed in Appendices~\ref{sec:annotation guideline} and~\ref{sec:annotator system ui}. We evaluate both the intermediate ``upstream'' components of our pipeline and the final ``downstream'' generated content. Furthermore, we investigate the feasibility of using an LLM-as-a-Judge as a proxy for human evaluation of the AI generated educational content by analyzing its agreement with our human experts.

\subsection{Upstream Component Quality}

First, we evaluate the quality of the upstream components that feed into the final generator: search query extraction and retrieved paper relevance. This evaluation was conducted on a set of 50 clinical cases. As shown in Table~\ref{tab:upstream_scores}, human experts found the search queries extracted by the system to be highly appropriate (Human Avg. score of 3.73). However, they were more critical of the relevance of the retrieved academic papers, giving an average score of 2.88. This suggests that while the system correctly identifies the main topics, the unfiltered, live-retrieved literature can contain articles that are not perfectly aligned with the specific clinical context of the report. In contrast, the LLM judges rated the paper relevance significantly higher (LLM Avg. 4.20), indicating a divergence in the assessment of contextual relevance between human experts and automated metrics.

\begin{table}[t!]
\centering
\resizebox{\columnwidth}{!}{%
\begin{tabular}{lcc}
\toprule
\textbf{Evaluator} & \textbf{Query} & \textbf{Paper} \\
& \textbf{Appropriateness} & \textbf{Relevance} \\
\midrule
\textbf{Human Evaluators} & \textbf{3.73} & \textbf{2.88} \\
\midrule
MedGemma-27B & 3.73 & 4.34 \\
GPT-4.1-mini & 4.15 & 4.52 \\
Gemini-2.5-Flash & 4.27 & 4.58 \\
Gemini-2.5-Pro & 4.03 & 3.37 \\
\cmidrule(l){2-3}
\textbf{LLM Avg.} & \textbf{4.05} & \textbf{4.20} \\
\bottomrule
\end{tabular}
}
\caption{Comparison of evaluator scores. The `` Evaluators'' row represents the combined results from two independent radiologists (n=50 each).}
\label{tab:upstream_scores}
\end{table}

\subsection{Downstream Generation Quality}
The primary evaluation focused on the quality of the final, user-facing outputs: textbook summaries, MCQs, and educational material. Three radiologists annotated the outputs from two generator models, Llama-3.3-70B-Instruct and MedGemma-27B\citep{sellergren2025medgemmatechnicalreport}. The detailed results are presented in Table~\ref{tab:human vs llm}.

\begin{table*}[t!]
\centering
\small 
\begin{tabular}{llcccc}
\toprule
\textbf{Model} & \textbf{Evaluator} & \begin{tabular}[c]{@{}c@{}}\textbf{Textbook} \\ \textbf{Summary}\end{tabular} & \begin{tabular}[c]{@{}c@{}}\textbf{Educational} \\ \textbf{Material}\end{tabular} & \begin{tabular}[c]{@{}c@{}}\textbf{MCQ} \\ \textbf{Quality}\end{tabular} & \begin{tabular}[c]{@{}c@{}}\textbf{Overall} \\ \textbf{Average}\end{tabular} \\
\midrule

\multirow{6}{*}{\textbf{Llama 3.3-70B-Instruct}} 
 & \textbf{Human Evaluators} & \textbf{3.43} & \textbf{3.78} & \textbf{3.11} & \textbf{3.44} \\
 \cmidrule(l){2-6}
 & MedGemma-27B & 3.64 & 3.66 & 3.79 & 3.70 \\
 & GPT-4.1-mini & 4.34 & 4.50 & 4.19 & 4.34 \\
 & Gemini-2.5-Flash & 2.82 & 3.58 & 4.08 & 3.49 \\
 & Gemini-2.5-Pro & 3.95 & 4.28 & 4.14 & 4.12 \\ 
 \cmidrule(l){2-6}
 & \textbf{LLM Avg.} & \textbf{3.69} & \textbf{4.01} & \textbf{4.05} & \textbf{3.91} \\
\midrule

\multirow{6}{*}{\textbf{MedGemma-27B}} 
 & \textbf{Human Evaluators} & \textbf{3.58} & \textbf{3.84} & \textbf{3.53} & \textbf{3.65} \\
 \cmidrule(l){2-6}
 & MedGemma-27B & 3.65 & 4.09 & 4.22 & 3.99 \\
 & GPT-4.1-mini & 4.21 & 4.79 & 4.60 & 4.53 \\
 & Gemini-2.5-Flash & 3.05 & 4.61 & 4.47 & 4.04 \\
 & Gemini-2.5-Pro & 3.84 & 4.18 & 4.15 & 4.06 \\
 \cmidrule(l){2-6}
 & \textbf{LLM Avg.} & \textbf{3.69} & \textbf{4.42} & \textbf{4.36} & \textbf{4.16} \\
\bottomrule
\end{tabular}
\caption{Main Generation Task Quality: Direct Comparison of Human Expert and LLM-as-a-Judge Evaluations. The ``Human Evaluators'' scores represent the combined results from three independent radiologists (n=50 each). All scores are on a 1-5 scale (5=best).}
\label{tab:human vs llm}
\end{table*}

Both models produced high-quality outputs according to our expert evaluators. Llama 3.3-70B-Instruct achieved a respectable average human score of 3.44, demonstrating its capability in synthesizing complex medical information into educational content. MedGemma-27B, a model more specialized for the medical domain, performed slightly better, with an average human score of 3.65. The experts particularly noted the higher quality of the MCQs generated by MedGemma-27B (3.53) compared to those from Llama-3.3-70B-Instruct (3.11). This suggests that the domain-specific nature of MedGemma-27B provides a distinct advantage in generating educational content, such as plausible distractors for multiple-choice questions.

When comparing human evaluations to the LLM-as-a-Judge scores, we note an interesting trend. The LLM judges also preferred MedGemma-27B over Llama 3.3, aligning with the relative ranking of the human experts. However, the LLMs consistently assigned higher absolute scores than the human radiologists. This suggests that while LLM-as-a-Judge can be a valuable tool for scalable, relative comparisons between models, its scoring calibration differs from that of human experts, indicating a tendency for score inflation. These findings suggest a promising path toward semi-automated evaluation while reinforcing the role of human experts as the gold standard for assessing clinical utility. Full LLM-as-a-Judge results are in Tab~\ref{sec:llm_as_judge_eval}.

\subsection{Inter-Annotator Agreement}

To ensure the reliability of our human evaluations, we measured the inter-annotator agreement (IAA) between the two board-certified radiologists using Krippendorff's Alpha \citep{krippendorff2011computing}.

The alpha coefficient is calculated as:
$\alpha = 1 - \frac{D_o}{D_e}$
Here, $D_o$ is the observed disagreement, calculated as the average difference between the ratings from each human annotator, $A_1$ and $A_2$, across all $M$ evaluated items. Specifically, if $r_{i,1}$ and $r_{i,2}$ are the ratings for item $i$ from $A_1$ and $A_2$ respectively, then:
$$
D_o = \frac{1}{M} \sum_{i=1}^{M} \delta^2(r_{i,1}, r_{i,2})
$$
$D_e$ represents the disagreement expected by chance, calculated based on the individual rating distributions of $A_1$ and $A_2$. For the difference function $\delta^2$, we first recoded the 1-to-5 Likert scale ratings into a 3-point interval scale (1-2 $\to$ 1; 3 $\to$ 2; 4-5 $\to$ 3) and then applied a squared difference:
$
\delta^2(u, v) = (u - v)^2
$. 

The results, presented in Table \ref{tab:iaa_detailed}, show a range of agreement levels. We observed good agreement for the \textit{Textbook Summary} from MedGemma-27B ($\alpha = 0.661$) and fair agreement for \textit{Paper Relevance} ($\alpha = 0.493$). 

Overall, our annotators demonstrated moderate to good agreement across most tasks (with the exception of MCQ quality), which is in line with agreement levels reported in prior work on high-quality datasets~\citep{liu2024benchmarkinggenerationevaluationcapabilities, bavaresco2025llmsinsteadhumanjudges}. The lower agreement for MCQ evaluation ($\alpha = 0.048$) suggests that the criteria for this specific task may require more detailed guidelines to improve consistency.

\section{Conclusion}

In this work, we introduce \textbf{\ours}, a novel, open-source system designed to augment clinical education by transforming clinical reports into structured, evidence-backed learning modules. Our system addresses the critical challenges of factual accuracy and knowledge freshness in medical AI by employing a sophisticated RAG pipeline. This pipeline features a hybrid retrieval mechanism that synthesizes knowledge from both foundational medical textbooks and real-time academic literature, ensuring the generated educational modules are both reliable and current.

Our rigorous evaluation, conducted by board-certified radiologists, confirmed that \ours{} can produce high-quality, clinically valuable educational content. Furthermore, our large-scale LLM-as-a-Judge analysis revealed a moderate but promising correlation with human expert judgments, suggesting a viable path toward scalable automated evaluation while underscoring the continued importance of expert oversight.

By publicly releasing the \ours{} system, its user interface, and the comprehensive evaluation dataset, we make two key contributions. First, we provide a practical tool that can be immediately adapted by other institutions to enhance their own training programs. Second, we offer a valuable benchmark and framework for future research into building trustworthy and effective generative AI systems for the high-stakes medical domain. We believe this work represents a significant step toward fostering more effective and efficient clinician-AI collaboration in medical education.

\section{Limitations}

While \ours demonstrates a promising approach to augmenting medical education, we acknowledge several limitations that offer avenues for future work.

First, our evaluation is primarily focused on the domain of radiology. Although the system's architecture is designed to be generalizable, its effectiveness and the nuances of its application in other medical specialties with different reporting styles and knowledge structures, such as pathology or cardiology, have not yet been explored. Future studies should assess the adaptability and performance of \ours across a broader range of clinical domains.

Second, the human evaluation, while rigorous and conducted by domain experts, was performed on a dataset of 50 clinical cases. A larger-scale study involving a greater number of cases and a more diverse cohort of radiologists would be beneficial to further validate our findings and provide more robust statistical power to the conclusions drawn.

Finally, our analysis of inter-annotator agreement and the LLM-as-a-Judge evaluations highlights challenges in consistently generating high-quality subjective content. The lower agreement scores for MCQs, for instance, suggest that these outputs require further refinement. This indicates that more advanced prompting techniques, fine-tuning of the generator models, or more sophisticated evaluation guidelines may be necessary to improve the reliability and educational value of these more complex, creative tasks.

\section*{Acknowledgments}

This research was supported by a grant of the Korea Health Technology R\&D Project through the Korea Health Industry Development Institute (KHIDI), funded by the Ministry of Health \& Welfare, Republic of Korea (grant number: HI19C1352).

\clearpage

\bibliography{custom}

\appendix
\onecolumn 

\newtcolorbox{reportbox}[1][]{
  colback=blue!5!white,
  colframe=blue!75!black,
  fonttitle=\bfseries,
  coltitle=white,
  colbacktitle=blue!75!black,
  title=Original Radiology Report,
  sharp corners,
  breakable,
  #1
}

\newtcolorbox{feedbackbox}[1][]{
  colback=green!5!white,
  colframe=ForestGreen,
  fonttitle=\bfseries,
  coltitle=white,
  colbacktitle=ForestGreen,
  title=Generated Educational Materials,
  sharp corners,
  breakable,
  #1
}

\newtcolorbox{infobox}[1][]{
  colback=black!5,      
  colframe=black!60,
  fonttitle=\bfseries,
  coltitle=black,
  sharp corners,
  #1
}

\newtcolorbox{promptbox}[1][]{
  colback=orange!5!white,  
  colframe=orange!80!black,
  fonttitle=\bfseries,
  coltitle=white,
  colbacktitle=orange!80!black, 
  title=System Prompt,
  sharp corners,
  breakable,
  #1
}


\startcontents[appendix]

\begin{center}
    \Large\bfseries Appendix Contents
\end{center}
\vspace{1em}

\begingroup
    \hypersetup{linkcolor=black}
    
    \printcontents[appendix]
        {l} 
        {1} 
        {\setcounter{tocdepth}{2}} 
\endgroup 

\clearpage

\section{LLM-as-a-Judge Evaluation}
\label{sec:llm_as_judge_eval}

\begin{table*}[hbt!]
\centering
\small
\begin{tabular}{llcccc}
\toprule
\textbf{Model} & \textbf{Judge} & \begin{tabular}[c]{@{}c@{}}\textbf{Textbook}\\ \textbf{Summary}\end{tabular} & \begin{tabular}[c]{@{}c@{}}\textbf{Educational}\\ \textbf{Material}\end{tabular} & \textbf{MCQ} & \textbf{Average} \\
\midrule

\multirow{5}{*}{\textbf{Llama 3.1-8B-Instruct}} 
    & MedGemma-27B & 3.64 ($\pm$0.95) & 3.49 ($\pm$1.10) & 3.69 ($\pm$1.05) & 3.61 \\
    & GPT-4.1-mini & 4.06 ($\pm$0.85) & 4.18 ($\pm$0.90) & 3.86 ($\pm$0.92) & 4.03 \\
    & Gemini-2.5-Pro & 3.59 ($\pm$1.15) & 3.55 ($\pm$1.20) & 3.92 ($\pm$1.18) & 3.69 \\
    & Gemini-2.5-Flash & 3.64 ($\pm$1.30) & 3.49 ($\pm$1.25) & 3.88 ($\pm$1.28) & 3.67 \\
\cmidrule(l){2-6}
    & \textbf{Avg. (Judges)} & \textbf{3.73} & \textbf{3.68} & \textbf{3.84} & \textbf{3.75} \\
\midrule

\multirow{5}{*}{\textbf{Qwen3-8B}} 
    & MedGemma-27B & 3.39 ($\pm$1.01) & 4.01 ($\pm$0.95) & 3.78 ($\pm$0.88) & 3.73 \\
    & GPT-4.1-mini & 3.42 ($\pm$0.90) & 4.49 ($\pm$0.75) & 4.22 ($\pm$0.81) & 4.04 \\
    & Gemini-2.5-Pro & 3.45 ($\pm$1.10) & 4.11 ($\pm$0.99) & 3.81 ($\pm$0.95) & 3.79 \\
    & Gemini-2.5-Flash & 3.39 ($\pm$1.25) & 4.01 ($\pm$1.15) & 3.75 ($\pm$1.05) & 3.72 \\
\cmidrule(l){2-6}
    & \textbf{Avg. (Judges)} & \textbf{3.41} & \textbf{4.16} & \textbf{3.89} & \textbf{3.82} \\
\midrule

\multirow{5}{*}{\textbf{Llama-4-Scout-17B-16E-Instruct}} 
    & MedGemma-27B & 3.68 ($\pm$0.88) & 4.08 ($\pm$0.85) & 3.85 ($\pm$0.80) & 3.87 \\
    & GPT-4.1-mini & 4.30 ($\pm$0.70) & 4.28 ($\pm$0.65) & 4.18 ($\pm$0.72) & 4.25 \\
    & Gemini-2.5-Pro & 3.71 ($\pm$0.95) & 4.15 ($\pm$0.90) & 4.01 ($\pm$0.88) & 3.96 \\
    & Gemini-2.5-Flash & 3.68 ($\pm$1.10) & 4.08 ($\pm$1.05) & 3.95 ($\pm$1.00) & 3.90 \\
\cmidrule(l){2-6}
    & \textbf{Avg. (Judges)} & \textbf{3.84} & \textbf{4.15} & \textbf{4.00} & \textbf{4.00} \\
\midrule

\multirow{5}{*}{\textbf{Qwen3-32B}} 
    & MedGemma-27B & 3.55 ($\pm$0.75) & 4.64 ($\pm$0.60) & 3.99 ($\pm$0.65) & 4.06 \\
    & GPT-4.1-mini & 3.99 ($\pm$0.65) & 4.19 ($\pm$0.50) & 4.48 ($\pm$0.55) & 4.22 \\
    & Gemini-2.5-Pro & 3.61 ($\pm$0.80) & 4.70 ($\pm$0.70) & 4.25 ($\pm$0.75) & 4.19 \\
    & Gemini-2.5-Flash & 3.55 ($\pm$1.20) & 4.64 ($\pm$1.10) & 4.18 ($\pm$1.00) & 4.12 \\
\cmidrule(l){2-6}
    & \textbf{Avg. (Judges)} & \textbf{3.68} & \textbf{4.54} & \textbf{4.23} & \textbf{4.15} \\
\midrule

\multirow{5}{*}{\textbf{Llama-3.3-70B-Instruct}} 
    & MedGemma-27B & 3.64 ($\pm$0.68) & 3.66 ($\pm$0.61) & 3.79 ($\pm$0.55) & 3.70 \\
    & GPT-4.1-mini & 4.34 ($\pm$0.72) & 4.50 ($\pm$0.55) & 4.19 ($\pm$0.60) & 4.34 \\
    & Gemini-2.5-Pro & 3.95 ($\pm$1.23) & 4.28 ($\pm$0.38) & 4.14 ($\pm$0.55) & 4.12 \\
    & Gemini-2.5-Flash & 2.82 ($\pm$1.45) & 3.58 ($\pm$1.44) & 4.08 ($\pm$1.46) & 3.49 \\
\cmidrule(l){2-6}
    & \textbf{Avg. (Judges)} & \textbf{3.69} & \textbf{4.01} & \textbf{4.05} & \textbf{3.91} \\
\midrule

\multirow{5}{*}{\textbf{MedGemma-27B}} 
    & MedGemma-27B & 3.65 ($\pm$0.88) & 4.09 ($\pm$0.61) & 4.22 ($\pm$0.60) & 3.99 \\
    & GPT-4.1-mini & 4.21 ($\pm$0.81) & 4.79 ($\pm$0.48) & 4.60 ($\pm$0.51) & 4.53 \\
    & Gemini-2.5-Pro & 3.84 ($\pm$1.18) & 4.18 ($\pm$0.45) & 4.15 ($\pm$0.72) & 4.06 \\
    & Gemini-2.5-Flash & 3.05 ($\pm$1.58) & 4.61 ($\pm$0.90) & 4.47 ($\pm$1.18) & 4.04 \\
\cmidrule(l){2-6}
    & \textbf{Avg. (Judges)} & \textbf{3.69} & \textbf{4.42} & \textbf{4.36} & \textbf{4.16} \\
\bottomrule
\end{tabular}
\caption{
    Aggregated LLM-as-a-Judge evaluation results across all datasets, comparing different judges.
    The \textbf{Avg. (Judges)} row indicates the mean of scores across the judges. All scores are on a 1-5 scale (5=best). Llama-4-Scout-17B-16E-Instruct\citep{meta_llama4} was inferenced in FP8.
}
\label{tab:multi_judge_aggregated_eval}
\end{table*}

\clearpage

\section{Detailed Inter-Annotator Agreement}
\label{sec:inter annotator agreement}

\begin{table*}[hbt!]
\centering
\small
\begin{tabular}{llccccc}
\toprule
\textbf{Model Context} & \textbf{Evaluation Metric} & \textbf{\begin{tabular}[c]{@{}c@{}}Krippendorff's \\ Alpha ($\alpha$)\end{tabular}} & \textbf{\begin{tabular}[c]{@{}c@{}}Pairwise \\ Kappa ($\kappa$)\end{tabular}} & \textbf{\begin{tabular}[c]{@{}c@{}}\% Exact \\ Agreement\end{tabular}} & \textbf{\begin{tabular}[c]{@{}c@{}}\% Within \\ $\pm$1 Point\end{tabular}} & \textbf{\begin{tabular}[c]{@{}c@{}}Correlation \\ (r)\end{tabular}} \\
\midrule

\multirow{2}{*}{Upstream} 
 & Keyword Appropriateness & 0.335 & 0.627 & 41\% & 80\% & 0.709 \\
 & Paper Relevance & 0.474 & 0.675 & 59\% & 95\% & 0.778 \\
\cmidrule(l){2-7}

\multirow{3}{*}{Llama3-70B} 
 & Textbook Summary & 0.347 & 0.555 & 48\% & 84\% & 0.587 \\
 & Educational Material & -0.228 & 0.382 & 50\% & 94\% & 0.325 \\
 & MCQ & -0.159 & 0.222 & 29\% & 81\% & 0.375 \\
\cmidrule(l){2-7}

\multirow{3}{*}{MedGemma-27B} 
 & Textbook Summary & 0.627 & 0.812 & 66\% & 96\% & 0.721 \\
 & Educational Material & 0.354 & 0.673 & 72\% & 100\% & 0.589 \\
 & MCQ & 0.114 & 0.629 & 46\% & 90\% & 0.596 \\
\bottomrule
\end{tabular}
\caption{Detailed Inter-Annotator Agreement (IAA) between three radiologists across different evaluation tasks. Krippendorff's Alpha ($\alpha$) and Avg. Pairwise Kappa ($\kappa$) measure reliability, while agreement percentages and Pearson correlation (r) provide further insight into rater consistency.}
\label{tab:iaa_detailed}
\end{table*}

Overall, the agreement scores suggest that MedGemma-27B's outputs were evaluated more consistently by the radiologists than those from Llama3.3-70B. As shown in Table \ref{tab:iaa_detailed}, MedGemma-27B's Textbook Summary achieved the highest reliability, with a Krippendorff's Alpha of 0.627, approaching the threshold for acceptable agreement, and a substantial average pairwise Kappa of 0.812. The upstream task of Paper Relevance also demonstrated moderate to substantial agreement across most measures.

\begin{figure*}[hbt!]
    \centering
    \includegraphics[width=0.8\linewidth]{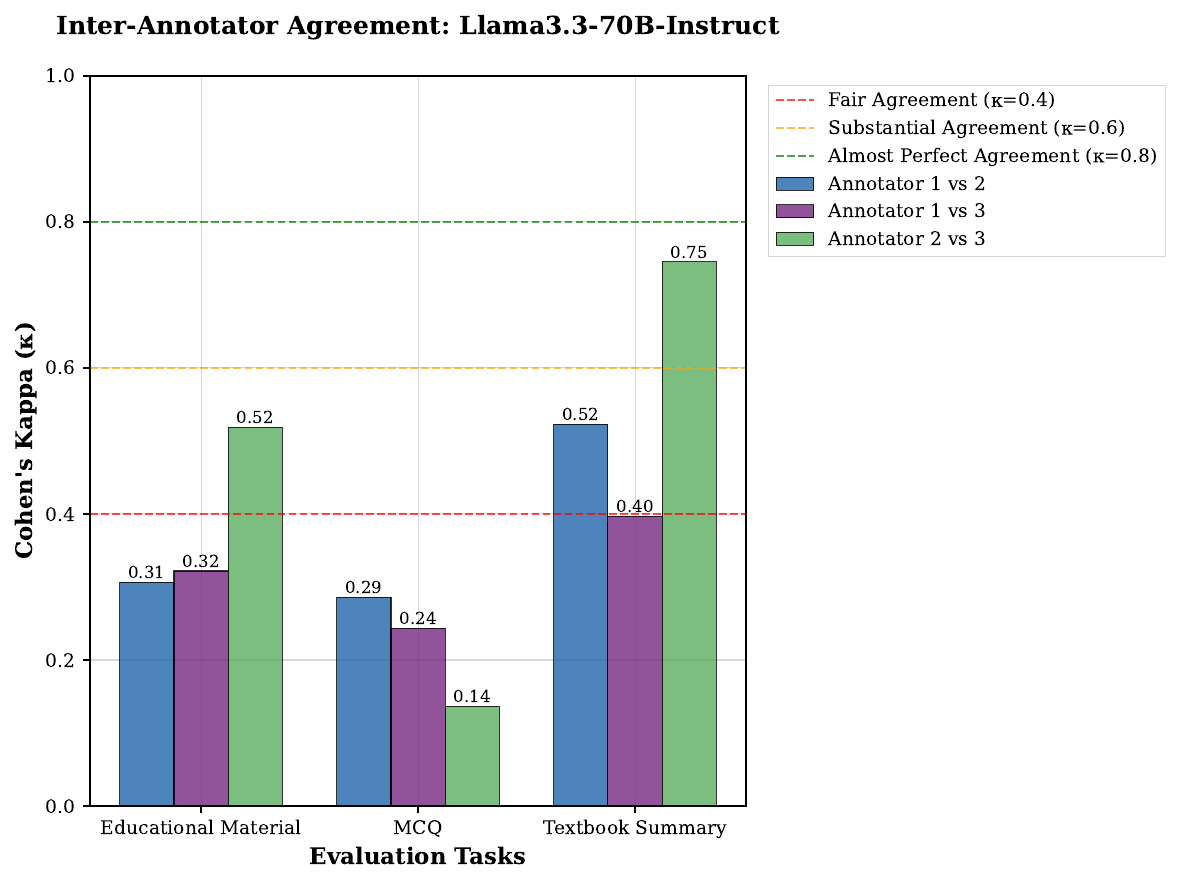}
    \caption{Pairwise Cohen's Kappa ($\kappa$) scores for Llama3.3-70B-Instruct Generated Content.}
    \label{fig:kappa_llama}
\end{figure*}

\begin{figure*}[hbt!]
    \centering
    \includegraphics[width=0.8\linewidth]{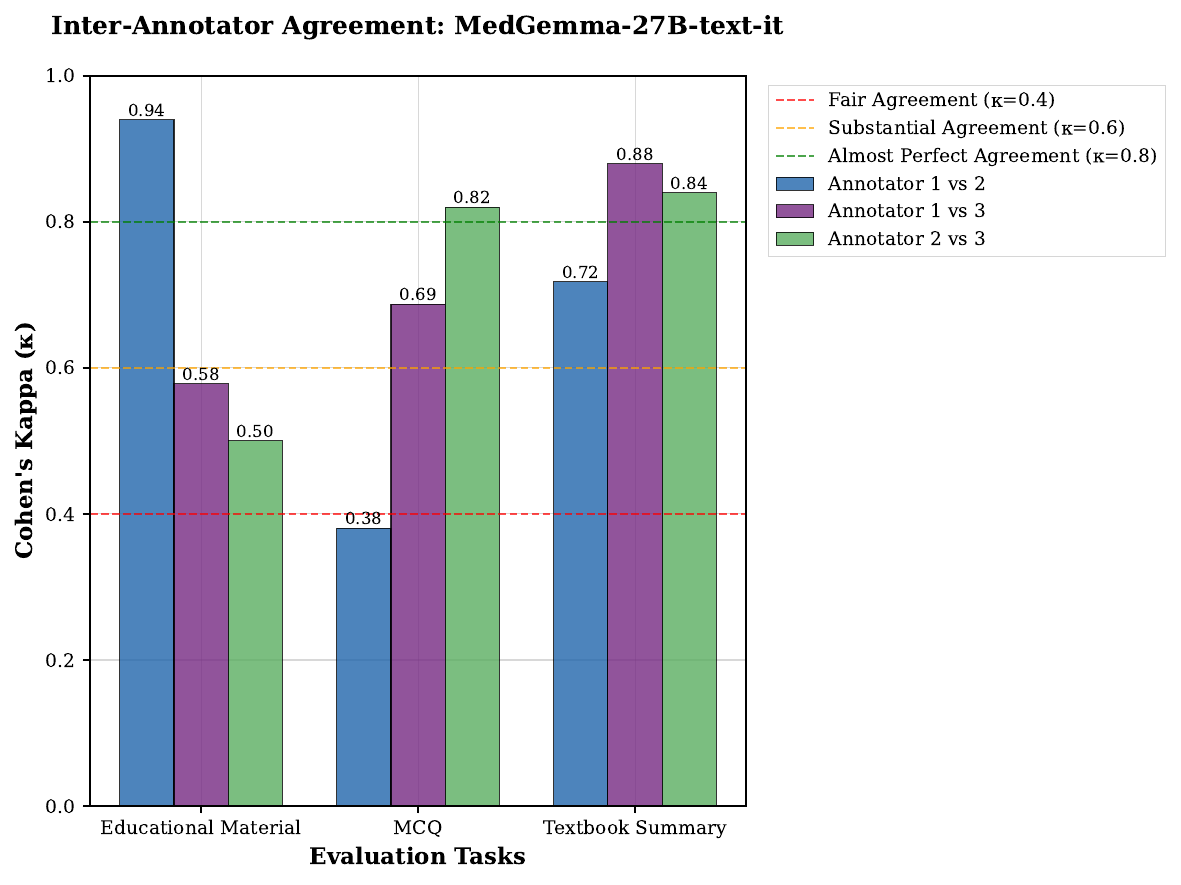}
    \caption{Pairwise Cohen's Kappa ($\kappa$) scores for MedGemma-27B-text-it Generated Content.}
    \label{fig:kappa_medgemma}
\end{figure*}

Conversely, the outputs from Llama3.3-70B, particularly for more subjective tasks like Educational Material and MCQ evaluation, yielded negative Alpha values, indicating systematic disagreement among the raters (Figure \ref{fig:kappa_llama}). The evaluation of MCQs proved challenging for both models, though agreement was notably higher for MedGemma-27B (Figure \ref{fig:kappa_medgemma}). These findings highlight that while structured summarization tasks can achieve high inter-rater reliability, evaluating more complex, subjective-generative tasks may require more detailed guidelines to ensure rater consistency.

\begin{figure*}[hbt!]
    \centering
    \includegraphics[width=0.8\linewidth]{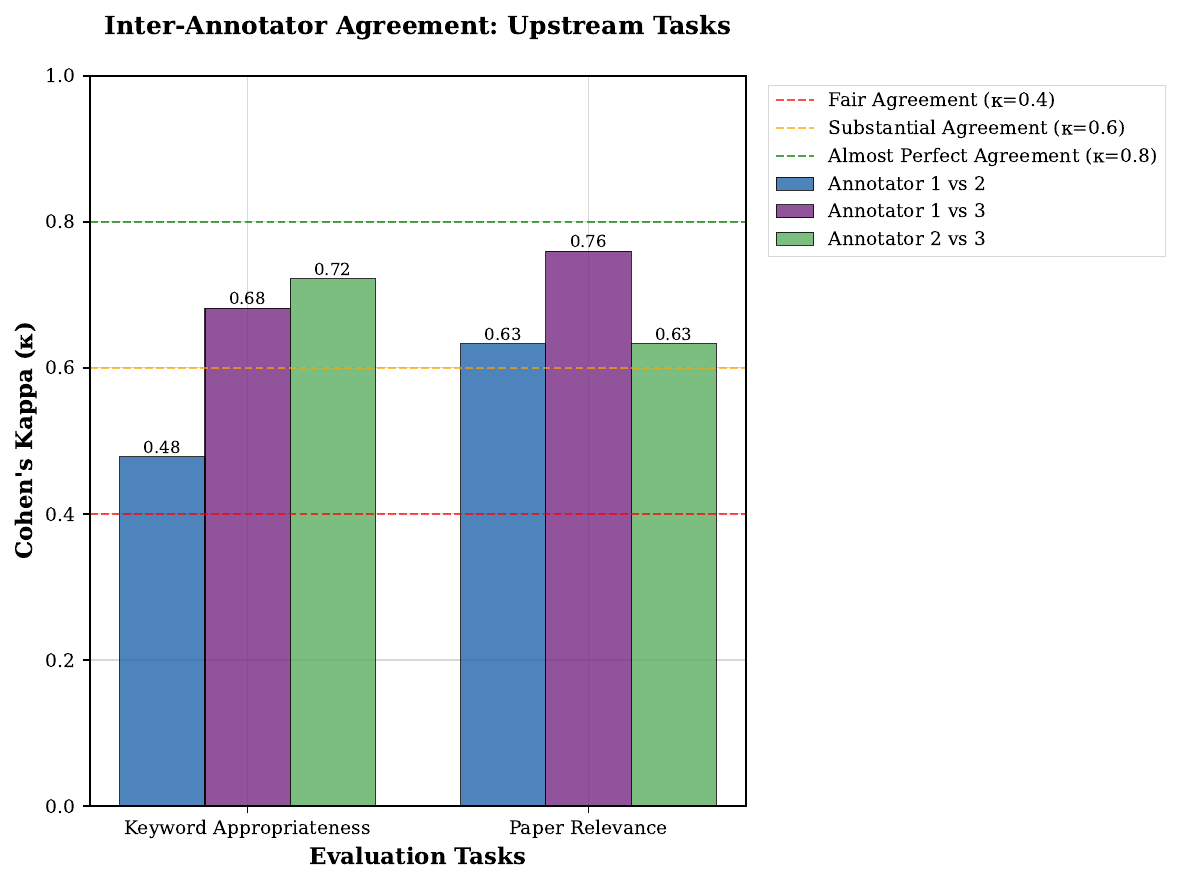}
    \caption{Pairwise Cohen's Kappa ($\kappa$) scores for Upstream Tasks. This figure shows the agreement between three pairs of annotators for keyword appropriateness and paper relevance.}
    \label{fig:kappa_upstream}
\end{figure*}

\clearpage

\section{MedTutor Dataset Samples and Public Dataset Information}
\label{sec:medtutor dataset samples}

This appendix provides the best inter-annotator agreement examples of the highest and lowest-scoring cases from the 50 cases evaluated by three expert radiologists. These cases were sampled (10 from each of the 5 datasets; Yale Internal, MIMIC-IV-note, MIMIC-CXR, CheXpert-Plus, ReXGradient-160K) and generated by two different models: Llama-3.3-70B-Instruct and MedGemma-27B-text-it. Each example includes the original clinical report and its corresponding generated report from \ours.

Also, we publicly release a large-scale \href{https://huggingface.co/datasets/yale-nlp/MedTutor}{dataset}(total 144K) generated by our system. This includes reports from \textbf{CheXpert-Plus}, \textbf{MIMIC-IV-note}, and \textbf{MIMIC-CXR} (2,000 reports each), processed by six different generator models and 4 evaluator models. Due to licensing and de-identification challenges, the Yale-internal and ReXGradient datasets are not included in the public release.

\begin{itemize}
    \item[\textbf{\hyperref[sec:case_c1]{C.1}}] Highest-scoring case generated by \texttt{Llama-3.3-70B-Instruct}.
    \item[\textbf{\hyperref[sec:case_c2]{C.2}}] Lowest-scoring case generated by \texttt{Llama-3.3-70B-Instruct}.
    \item[\textbf{\hyperref[sec:case_c3]{C.3}}] Highest-scoring case generated by \texttt{MedGemma-27B-text-it}.
    \item[\textbf{\hyperref[sec:case_c4]{C.4}}] Lowest-scoring case generated by \texttt{MedGemma-27B-text-it}.
\end{itemize}

\newpage
\subsection{Highest-Scoring Case with Llama-3.3-70B-Instruct}
\label{sec:case_c1}

\begin{infobox}[title=Case Information]
    \textbf{Dataset:} \texttt{MIMIC-IV-note} \\
    \textbf{Generator Model:} \texttt{Llama-3.3-70B-Instruct} \\
    \textbf{Case ID:} \url{19287224-RR-6}
\end{infobox}

\begin{reportbox}
    INDICATION:  NO\_PO contrast; History: () with abd pain NO\_PO contrast// abd \par pain r/o appendicitiis \par  \par TECHNIQUE:  Single phase split bolus contrast: MDCT axial images were acquired \par through the abdomen and pelvis following intravenous contrast administration \par with split bolus technique. \par Oral contrast was administered. \par Coronal and sagittal reformations were performed and reviewed on PACS. \par  \par DOSE:  Acquisition sequence: \par    1) Stationary Acquisition 7.5 s, 0.5 cm; CTDIvol = 35.2 mGy (Body) DLP = \par 17.6 mGy-cm. \par    2) Spiral Acquisition 7.3 s, 55.8 cm; CTDIvol = 9.8 mGy (Body) DLP = 548.4 \par mGy-cm. \par  Total DLP (Body) = 566 mGy-cm. \par  \par COMPARISON:  None. \par  \par FINDINGS:  \par  \par LOWER CHEST: Visualized lung fields are within normal limits. There is no \par evidence of pleural or pericardial effusion. \par  \par ABDOMEN: \par  \par HEPATOBILIARY: The liver demonstrates homogenous attenuation throughout.  \par There is no evidence of focal lesions.  There is no evidence of intrahepatic \par or extrahepatic biliary dilatation.  The gallbladder is within normal limits. \par  \par PANCREAS: The pancreas has normal attenuation throughout, without evidence of \par focal lesions or pancreatic ductal dilatation.  There is no peripancreatic \par stranding. \par  \par SPLEEN: The spleen shows normal size and attenuation throughout, without \par evidence of focal lesions. \par  \par ADRENALS: The right and left adrenal glands are normal in size and shape. \par  \par URINARY: The kidneys are of normal and symmetric size with normal nephrogram.  \par There is no evidence of focal renal lesions or hydronephrosis.  There is no \par perinephric abnormality. \par  \par GASTROINTESTINAL: The stomach is unremarkable.  Small bowel loops demonstrate \par normal caliber, wall thickness, and enhancement throughout.  There are \par scattered colonic diverticula.  The appendix is fluid-filled and dilated with \par wall hyperemia.  A measures 13 mm with periappendiceal stranding and a small \par appendicular at the base.  There is no periappendiceal abscess. \par  \par PELVIS: The urinary bladder and distal ureters are unremarkable.  There is a \par small amount of free fluid in the pelvis. \par  \par REPRODUCTIVE ORGANS: The visualized reproductive organs are unremarkable. \par  \par LYMPH NODES: There is no retroperitoneal or mesenteric lymphadenopathy.  There \par is no pelvic or inguinal lymphadenopathy. \par  \par VASCULAR: There is no abdominal aortic aneurysm.  No significant \par atherosclerotic disease is noted. \par  \par BONES: There is no evidence of worrisome osseous lesions or acute fracture.  \par Moderate multilevel degenerative changes noted of the lumbar spine. \par  \par SOFT TISSUES: The abdominal and pelvic wall is within normal limits. \par  \par IMPRESSION: \par  \par  \par 1. Acute appendicitis. \par 2. Colonic diverticulosis without diverticulitis. \par  \par NOTIFICATION:  Findings wet read to the electronic dashboard and discussed \par with () \par 
\end{reportbox}

\begin{feedbackbox}
    \begin{longtable}{@{} p{\linewidth} @{}}
\textbf{Keywords:} Acute appendicitis, Colonic diverticulosis \\
\midrule
\textbf{Retrieved Paper 1:} \textbf{Retrospective case-control study to predict a potential underlying appendiceal tumor in an acute appendicitis context based on a CT-scoring system.} \par
OBJECTIVES: To assess CT signs to discriminate an appendiceal tumor versus a non-tumoral appendix in an acute appendicitis context. \par METHODS: A 10-year bicentric retrospective case-control study was performed in adults. Patients with a histopathological appendiceal tumor and appendicitis were paired for age and sex with patients with non-tumorous appendicitis (1/3 ratio, respectively). Two senior radiologists blindly analyzed numerous CT findings... (Omitted) \par
\small URL: https://pubmed.ncbi.nlm.nih.gov/33454458/ | Source: PubMed
\\
\midrule
\textbf{Retrieved Paper 2:} \textbf{Dual energy CT in acute appendicitis: value of low mono-energy.} \par
OBJECTIVES: To assess the potential role of low monoenergetic images in the evaluation of acute appendicitis. \par METHODS: A retrospective study of 42 patients with pathology proven acute appendicitis underwent contrast-enhanced-CT conducted on a single-source-DECT before surgery. Attenuation, SNR, and CNR were calculated on both monoenergetic and conventional images and compared to 24 abdominal CT-scans with normal appendix. Representative... (Omitted) \par
\small URL: https://pubmed.ncbi.nlm.nih.gov/33992882/ | Source: PubMed
\\
\midrule
\textbf{Retrieved Paper 3:} \textbf{Association between colorectal cancer and colonic diverticulosis: case-control study based on computed tomographic colonography.} \par
BACKGROUND: The association between colorectal cancer (CRC) and diverticulosis is uncertain, and there is little data reported in Asia. This study examined the association of both diseases based on the computed tomographic colonography (CTC). \par METHODS: This study was designed as a case-control study. Korean subjects undergoing CTC between April 2002 to April 2008 in Samsung Medical Center were eligible. Patients with histologically proven... (Omitted) \par
\small URL: https://pubmed.ncbi.nlm.nih.gov/21516446/ | Source: PubMed
\\
\midrule
\textbf{Retrieved Paper 4:} \textbf{REVISED CONCEPTS ON DIVERTICULAR DISEASE OF THE COLON. II. SO-CALLED DIVERTICULITIS: DIVERTICULAR SIGMOIDITIS AND PERISIGMOIDITIS; DIVERTICULAR ABSCESS, FISTULA, AND FRANK PERITONITIS.} \par
No Abstract \par
\small URL: https://pubmed.ncbi.nlm.nih.gov/14272097/ | Source: PubMed
\\
\midrule
\textbf{Textbook Summary 1} \par
Acute appendicitis is inflammation typically caused by luminal obstruction (fecolith, lymphoid hyperplasia) leading to ischemia. Radiologic findings on plain films are infrequent (<50\%) but may include an appendicolith, cecal wall thickening, or signs of perforation like extraluminal gas or loss of fat planes. CT is highly accurate, showing a distended appendix (>7mm), wall thickening, periappendicular fat stranding, and potentially an appendicolith or complications like perforation. Nonvisualization of the appendix on CT in the setting of right lower quadrant pain makes appendicitis highly unlikely.
\\
\midrule
\textbf{Textbook Summary 2} \par
Colonic diverticulosis involves acquired herniations of the colonic mucosa and submucosa through the muscularis propria, typically appearing as thin-walled, air-filled outpouchings on imaging. Radiographically, these diverticula are often seen as rounded structures, sometimes with a bubbly appearance, and may show residual contrast from prior studies. Ultrasound can show thin-walled outpouchings with echogenic material and potential shadowing, while CT demonstrates rounded outpouchings containing air or contrast, often associated with circumferential colonic wall thickening due to muscular hypertrophy.
\\
\midrule
\textbf{Assessment Question 1} \par
\textit{According to the provided textbook summary, which CT finding is highly suggestive of acute appendicitis?} \par
\begin{itemize}
    \item A. Normal appendix diameter
    \item B. Appendix diameter > 7mm
    \item C. Absence of periappendicular fat stranding
    \item D. Nonvisualization of the appendix
\end{itemize}
\textbf{Answer:} B \par
\textbf{Explanation:} The textbook summary states that CT findings suggestive of acute appendicitis include a distended appendix (>7mm).
\\
\midrule
\textbf{Assessment Question 2} \par
\textit{Based on the provided abstracts, which CT finding is a positive predictive factor for an underlying appendiceal neoplasm in the context of acute appendicitis?} \par
\begin{itemize}
    \item A. Appendicolith
    \item B. Fat stranding
    \item C. Mural calcifications
    \item D. Appendix diameter < 15mm
\end{itemize}
\textbf{Answer:} C \par
\textbf{Explanation:} The abstract from the study on appendiceal tumors identifies mural calcifications as a positive predictive factor (OR = 47, p = 0.0001) for an underlying tumor in patients presenting with acute appendicitis.
\\
\midrule
\textbf{Assessment Question 3} \par
\textit{What is the characteristic radiographic appearance of colonic diverticula as described in the textbook summary?} \par
\begin{itemize}
    \item A. Thick-walled outpouchings filled with fluid
    \item B. Thin-walled outpouchings, often rounded, containing air or contrast
    \item C. Irregular thickening of the colonic wall
    \item D. Absence of air within the colon
\end{itemize}
\textbf{Answer:} B \par
\textbf{Explanation:} The textbook summary describes colonic diverticula as acquired herniations appearing as thin-walled, air-filled outpouchings on imaging, often rounded and potentially containing contrast.
\\
\midrule
\textbf{Assessment Question 4} \par
\textit{According to the provided abstract analyzing the association between colorectal cancer and diverticulosis, what was the main conclusion regarding this association?} \par
\begin{itemize}
    \item A. Colorectal cancer is strongly associated with diverticulosis.
    \item B. Diverticulosis is more common in patients with right-sided colon cancer.
    \item C. Colorectal cancer is not associated with diverticulosis.
    \item D. Computed tomographic colonography is not useful for assessing diverticulosis.
\end{itemize}
\textbf{Answer:} C \par
\textbf{Explanation:} The abstract from the case-control study using CT colonography concluded that colorectal cancer is not associated with diverticulosis, as the distribution of diverticulosis was similar between the cancer and control groups.
\\
\midrule
\textbf{Final Educational Material} \par
\par\textbf{Radiology Feedback}\par\par\textbf{\# Acute Appendicitis}\par \par \textbullet~~Imaging Findings: CT is the preferred modality. Key findings include an appendix diameter >7 mm, wall thickening, periappendiceal fat stranding, and potentially an appendicolith. \par \textbullet~~Differential Diagnosis: Nonvisualization of the appendix on CT in the setting of right lower quadrant pain makes appendicitis highly unlikely. \par \textbullet~~Advanced Techniques: Low monoenergetic CT images can enhance visualization of inflamed bowel wall and mucosal enhancement, potentially improving diagnostic accuracy. \par \textbullet~~Clinical Correlation: While appendicitis is often caused by obstruction, CT can help identify potential underlying neoplasms, especially in atypical presentations. Features suggesting neoplasm include mural calcifications, an appendix mass, focal asymmetric wall abnormality, or diameter $\geq$ 15 mm. \par \par\textbf{\# Colonic Diverticulosis}\par \par \textbullet~~Imaging Findings: Diverticula appear as thin-walled, air-filled or contrast-filled outpouchings from the colon, typically rounded. CT demonstrates these outpouchings, often associated with circumferential colonic wall thickening due to muscular hypertrophy. \par \textbullet~~Clinical Correlation: Diverticulosis itself is the presence of these outpouchings and is generally asymptomatic. It is not directly associated with an increased risk of colorectal cancer.
\\
\end{longtable}
\end{feedbackbox}


\clearpage

\subsection{Lowest-Scoring Case with Llama-3.3-70B-Instruct}
\label{sec:case_c2}

\begin{infobox}[title=Case Information]
    \textbf{Dataset:} \texttt{RexGradient-160K} \\
    \textbf{Generator Model:} \texttt{Llama-3.3-70B-Instruct} \\
    \textbf{Case ID:} \url{pGRDNRZB2HZ56F0ZL\_aGRDNZSUP68GUCQN2\_s1.2.826.0.1.3680043.8.498.16841148712021794680399240737673267273}
\end{infobox}

\begin{reportbox}
    Indication: Fall yesterday. Fever this morning. \par  \par Findings: 4228 hours. The heart size and mediastinal contours are stable. There is stable mild subsegmental atelectasis at both lung bases. No confluent airspace opacity, pleural effusion or pneumothorax. Fracture of the proximal right humerus again noted. \par  \par Impression: Stable mild bibasilar atelectasis. No acute cardiopulmonary process.
\end{reportbox}

\begin{feedbackbox}
    \begin{longtable}{@{} p{\linewidth} @{}}
\textbf{Keywords:} atelectasis, fracture \\
\midrule
\textbf{Retrieved Paper 1:} \textbf{Debunking a mythology: Atelectasis is not a cause of postoperative fever.} \par
Most physicians appreciate that practicing medicine is a commitment to continuous learning. However, "learning" can be mistakenly understood as simply the acquisition of facts and new knowledge. But learning also necessitates the constant re-examination and challenging of one's existing body of knowledge, as misinformation persists when one's beliefs are not challenged or questioned in the light of new information. One example is the pervasive... (Omitted) \par
\small URL: https://pubmed.ncbi.nlm.nih.gov/39566396/ | Source: PubMed
\\
\midrule
\textbf{Retrieved Paper 2:} \textbf{Use of artificial intelligence in triaging of chest radiographs to reduce radiologists' workload.} \par
OBJECTIVES: To evaluate whether deep learning-based detection algorithms (DLD)-based triaging can reduce outpatient chest radiograph interpretation workload while maintaining noninferior sensitivity. \par METHODS: This retrospective study included patients who underwent initial chest radiography at the outpatient clinic between June 1 and June 30, 2017. Readers interpreted radiographs with/without a commercially available DLD that detects nine... (Omitted) \par
\small URL: https://pubmed.ncbi.nlm.nih.gov/37615766/ | Source: PubMed
\\
\midrule
\textbf{Retrieved Paper 3:} \textbf{Assessment of proximal tibial fractures with 3D FRACTURE (fast field echo resembling a CT using restricted echo-spacing) MRI-intra-individual comparison with CT.} \par
OBJECTIVES: To evaluate the feasibility and diagnostic performance of a 3D FRACTURE (fast field echo resembling a CT using restricted echo-spacing) MRI sequence for the detection and classification of proximal tibial fractures compared with CT. \par METHODS: We retrospectively included 126 patients (85 male; 39.6±14.5 years) from two centers following acute knee injury. Patients underwent knee MRI at 3T including FRACTURE-MRI. Additional CT was... (Omitted) \par
\small URL: https://pubmed.ncbi.nlm.nih.gov/40126605/ | Source: PubMed
\\
\midrule
\textbf{Retrieved Paper 4:} \textbf{How I Do It: Evaluating Cardiac Implantable Devices and Noncardiac Mimics on Chest Radiographs.} \par
Cardiac implantable electronic devices (CIEDs), including pacemakers and defibrillators, are increasingly used to manage various cardiac conditions. This article reviews the radiographic appearance, typical components, and placement of CIEDs, including newer technologies like leadless pacemakers and MRI-conditional devices. The article also highlights the imaging findings of common complications such as lead dislodgement, fracture, and... (Omitted) \par
\small URL: https://pubmed.ncbi.nlm.nih.gov/40358448/ | Source: PubMed
\\
\midrule
\textbf{Textbook Summary 1} \par
Atelectasis, or lung collapse, presents radiologically with increased lung density, vessel crowding, and potential fissure/mediastinal displacement. Obstructive atelectasis involves air resorption distal to a blockage (tumor, mucus plug, foreign body), while nonobstructive atelectasis retains some air. Other forms include passive (pleural effusion/pneumothorax), adhesive (surfactant deficiency), cicatrizing (fibrosis), and discoid/rounded atelectasis (often related to pleural inflammation or obstruction). Specific patterns, like the "Luftsichel" sign, can indicate left upper lobe collapse.
\\
\midrule
\textbf{Textbook Summary 2} \par
Enteropathy-associated T-cell lymphoma, a type of non-Hodgkin lymphoma, often presents with bowel wall thickening, ulceration, or strictures, particularly in the proximal small bowel. Radiologic findings may include circumferential wall thickening, mesenteric fat infiltration, and nonbulky lymphadenopathy, with a high frequency of FDG uptake on PET scans. Complications like bowel perforation are common, especially in Type II lymphoma, and differentiating it from large B-cell lymphoma or refractory celiac disease is crucial.
\\
\midrule
\textbf{Assessment Question 1} \par
\textit{According to the provided abstract debunking a mythology, what is the common misconception regarding atelectasis?} \par
\begin{itemize}
    \item A. Atelectasis is a common cause of postoperative consolidation.
    \item B. Atelectasis is a common cause of postoperative fever.
    \item C. Atelectasis is difficult to differentiate from pneumonia on imaging.
    \item D. Atelectasis is primarily caused by pleural effusion.
\end{itemize}
\textbf{Answer:} B \par
\textbf{Explanation:} The abstract explicitly states that the pervasive belief that postoperative atelectasis causes fever is a myth, despite ample evidence to the contrary.
\\
\midrule
\textbf{Assessment Question 2} \par
\textit{Which of the following is NOT a type of atelectasis mentioned in the textbook summary?} \par
\begin{itemize}
    \item A. Obstructive atelectasis
    \item B. Passive atelectasis
    \item C. Adhesive atelectasis
    \item D. Vascular atelectasis
\end{itemize}
\textbf{Answer:} D \par
\textbf{Explanation:} The textbook summary lists obstructive, nonobstructive, passive, adhesive, cicatrizing, and discoid/rounded atelectasis, but not vascular atelectasis.
\\
\midrule
\textbf{Assessment Question 3} \par
\textit{According to the abstract comparing FRACTURE-MRI and CT for proximal tibial fractures, what was the sensitivity of FRACTURE-MRI for detecting fractures?} \par
\begin{itemize}
    \item A. 78.2%
    \item B. 85.7%
    \item C. 91.5%
    \item D. 97.1%
\end{itemize}
\textbf{Answer:} C \par
\textbf{Explanation:} The abstract states that fractures were detected by FRACTURE-MRI with a sensitivity of 91.5
\\
\midrule
\textbf{Assessment Question 4} \par
\textit{The abstract on evaluating cardiac implantable devices mentions which potential complication related to device leads?} \par
\begin{itemize}
    \item A. Dislodgement
    \item B. Calcification
    \item C. Consolidation
    \item D. Cardiomegaly
\end{itemize}
\textbf{Answer:} A \par
\textbf{Explanation:} The abstract explicitly lists lead dislodgement, fracture, and perforation as common complications of cardiac implantable electronic devices.
\\
\midrule
\textbf{Final Educational Material} \par
Okay, here is the synthesized feedback focusing on atelectasis and fracture, based on the provided information: \par \par\textbf{\# Atelectasis}\par \par \textbullet~~Clinical Context: Atelectasis is commonly seen postoperatively and in the context of fever, but it is not a direct cause of fever itself. This is a persistent misconception. \par \textbullet~~Imaging Appearance: Radiographically, atelectasis appears as increased lung density, often with crowding of pulmonary vessels. It can be subsegmental, as seen in this case. \par \textbullet~~Differentiation: It's important to differentiate atelectasis from consolidation (infection), which typically presents with air bronchograms and may have a more lobar distribution. \par \textbullet~~AI Application: AI-based triaging tools can help identify atelectasis on chest radiographs, potentially reducing radiologist workload while maintaining diagnostic sensitivity. \par \par\textbf{\# Fracture}\par \par \textbullet~~Imaging Modalities: While CT is the standard for fracture evaluation, specialized MRI sequences like FRACTURE-MRI are being investigated as alternatives, potentially reducing radiation exposure and costs, especially when MRI is already indicated. \par \textbullet~~MRI Performance: Studies show FRACTURE-MRI has high sensitivity and specificity for detecting and classifying fractures, comparable to CT in certain locations like the proximal tibia. \par \textbullet~~Complications: Imaging is crucial for identifying complications of implanted devices, such as lead fracture in pacemakers/defibrillators, which can be visualized on chest radiographs.
\\
\end{longtable}
\end{feedbackbox}


\clearpage

\subsection{Highest-Scoring Case with MedGemma-27B-text-it}
\label{sec:case_c3}

\begin{infobox}[title=Case Information]
    \textbf{Dataset:} \texttt{CheXpert-Plus} \\
    \textbf{Generator Model:} \texttt{MedGemma-27B-text-it} \\
    \textbf{Case ID:} \url{23803}
\end{infobox}

\begin{reportbox}
    Findings:   Two views of the chest demonstrate reticular opacities  \par bilaterally, right greater than left, with a basilar predominance,  \par likely secondary to underlying fibrotic lung disease. There is no  \par evidence of pulmonary edema or pleural effusion. There is  \par cardiomegaly. Left anterior chest wall cardiac pacer appears  \par unchanged in position, with two ventricular leads.   \par   \par   \par  \par Impression:    \par   \par  1.  FINDINGS CONSISTENT WITH FIBROTIC LUNG DISEASE AS DESCRIBED  \par ABOVE.   \par   \par  2.  NO EVIDENCE OF PULMONARY EDEMA.   \par   \par  3.  STABLE CARDIOMEGALY WITH STABLE ARRANGEMENT OF PACER LEADS.   \par   \par  
\end{reportbox}

\begin{feedbackbox}
    \begin{longtable}{@{} p{\linewidth} @{}}
\textbf{Keywords:} Fibrotic Lung Disease \\
\midrule
\textbf{Retrieved Paper 1:} \textbf{Collaborative radiologic and histopathologic assessment of fibrotic lung disease.} \par
The idiopathic interstitial pneumonias (IIPs) are a seemingly disconnected collection of diseases usually associated with the presence of pulmonary fibrosis. Categorization of the IIPs continues to be problematic despite recent attempts to refine the diagnostic criteria and suggests that rather than separate diseases, these pneumonias represent a spectrum of injury and abnormal repair of the alveolar wall. Although the initiating injury or... (Omitted) \par
\small URL: https://pubmed.ncbi.nlm.nih.gov/20501710/ | Source: PubMed
\\
\midrule
\textbf{Retrieved Paper 2:} \textbf{From the radiologic pathology archives: organization and fibrosis as a response to lung injury in diffuse alveolar damage, organizing pneumonia, and acute fibrinous and organizing pneumonia.} \par
Organization, characterized by fibroblast proliferation, is a common and nearly universal response to lung injury whether it is focal or diffuse. Despite the vast range of injurious agents, the lung's response to injury is quite limited, with a similar pattern of reaction seen radiologically and histologically regardless of the underlying cause. Although there is a tendency to divide organization into distinct entities, the underlying injury to... (Omitted) \par
\small URL: https://pubmed.ncbi.nlm.nih.gov/24224590/ | Source: PubMed
\\
\midrule
\textbf{Textbook Summary 1} \par
In fibrotic lung disease, key radiologic findings include a reticulonodular pattern, ground-glass opacities, and honeycombing, with a basilar and subpleural distribution. Diagnostic considerations for fibrotic lung disease, such as usual interstitial pneumonia (UIP), involve identifying a combination of clinical and imaging features, including progressive dyspnea, clubbing, and bibasilar crackles, along with characteristic HRCT findings like traction bronchiectasis and macrocystic honeycombing. HRCT plays a crucial role in differentiating UIP from other idiopathic interstitial pneumonias and guiding biopsy sites for definitive diagnosis.
\\
\midrule
\textbf{Assessment Question 1} \par
\textit{What is a characteristic radiologic finding in fibrotic lung disease?} \par
\begin{itemize}
    \item A. Pulmonary edema
    \item B. Reticulonodular pattern
    \item C. Cardiomegaly
    \item D. Pleural effusion
\end{itemize}
\textbf{Answer:} B \par
\textbf{Explanation:} Fibrotic lung disease is characterized by specific radiologic findings, including a reticulonodular pattern, which is a key feature of this condition.
\\
\midrule
\textbf{Assessment Question 2} \par
\textit{What is the typical distribution of radiologic findings in fibrotic lung disease?} \par
\begin{itemize}
    \item A. Apical and central
    \item B. Basilar and subpleural
    \item C. Upper lobe and peripheral
    \item D. Diffuse and homogeneous
\end{itemize}
\textbf{Answer:} B \par
\textbf{Explanation:} Fibrotic lung disease typically shows a basilar and subpleural distribution of radiologic findings, which is an important characteristic for diagnosis.
\\
\midrule
\textbf{Final Educational Material} \par
\par\textbf{Fibrotic Lung Disease}\par \par Fibrotic lung disease is characterized by key radiologic findings including a reticulonodular pattern, ground-glass opacities, and honeycombing, typically with a basilar and subpleural distribution. The diagnosis often involves identifying a combination of clinical features such as progressive dyspnea, clubbing, and bibasilar crackles, along with characteristic HRCT findings like traction bronchiectasis and macrocystic honeycombing. High-resolution computed tomography (HRCT) plays a crucial role in differentiating usual interstitial pneumonia (UIP) from other idiopathic interstitial pneumonias and in guiding biopsy sites for a definitive diagnosis. A collaborative diagnostic approach combining radiologic and histopathologic assessments can lead to a more reliable identification of the predominant pathways leading to pulmonary fibrosis, which has implications for therapy and future research directions.
\\
\end{longtable}
\end{feedbackbox}


\clearpage

\subsection{Lowest-Scoring Case with MedGemma-27B-text-it}
\label{sec:case_c4}

\begin{infobox}[title=Case Information]
    \textbf{Dataset:} \texttt{RexGradient-160K} \\
    \textbf{Generator Model:} \texttt{MedGemma-27B-text-it} \\
    \textbf{Case ID:} \url{pGRDNRZB2HZ56F0ZL_aGRDNZSUP68GUCQN2_s1.2.826.0.1.3680043.8.498.16841148712021794680399240737673267273}
\end{infobox}

\begin{reportbox}
    Indication: Fall yesterday. Fever this morning. \par  \par Findings: 4228 hours. The heart size and mediastinal contours are stable. There is stable mild subsegmental atelectasis at both lung bases. No confluent airspace opacity, pleural effusion or pneumothorax. Fracture of the proximal right humerus again noted. \par  \par Impression: Stable mild bibasilar atelectasis. No acute cardiopulmonary process.
\end{reportbox}

\begin{feedbackbox}
    \begin{longtable}{@{} p{\linewidth} @{}}
\textbf{Keywords:} atelectasis, fracture \\
\midrule
\textbf{Retrieved Paper 1:} \textbf{Debunking a mythology: Atelectasis is not a cause of postoperative fever.} \par
Most physicians appreciate that practicing medicine is a commitment to continuous learning. However, "learning" can be mistakenly understood as simply the acquisition of facts and new knowledge. But learning also necessitates the constant re-examination and challenging of one's existing body of knowledge, as misinformation persists when one's beliefs are not challenged or questioned in the light of new information. One example is the pervasive... (Omitted) \par
\small URL: https://pubmed.ncbi.nlm.nih.gov/39566396/ | Source: PubMed
\\
\midrule
\textbf{Retrieved Paper 2:} \textbf{Use of artificial intelligence in triaging of chest radiographs to reduce radiologists' workload.} \par
OBJECTIVES: To evaluate whether deep learning-based detection algorithms (DLD)-based triaging can reduce outpatient chest radiograph interpretation workload while maintaining noninferior sensitivity. \par METHODS: This retrospective study included patients who underwent initial chest radiography at the outpatient clinic between June 1 and June 30, 2017. Readers interpreted radiographs with/without a commercially available DLD that detects nine... (Omitted) \par
\small URL: https://pubmed.ncbi.nlm.nih.gov/37615766/ | Source: PubMed
\\
\midrule
\textbf{Retrieved Paper 3:} \textbf{Assessment of proximal tibial fractures with 3D FRACTURE (fast field echo resembling a CT using restricted echo-spacing) MRI-intra-individual comparison with CT.} \par
OBJECTIVES: To evaluate the feasibility and diagnostic performance of a 3D FRACTURE (fast field echo resembling a CT using restricted echo-spacing) MRI sequence for the detection and classification of proximal tibial fractures compared with CT. \par METHODS: We retrospectively included 126 patients (85 male; 39.6±14.5 years) from two centers following acute knee injury. Patients underwent knee MRI at 3T including FRACTURE-MRI. Additional CT was... (Omitted) \par
\small URL: https://pubmed.ncbi.nlm.nih.gov/40126605/ | Source: PubMed
\\
\midrule
\textbf{Retrieved Paper 4:} \textbf{How I Do It: Evaluating Cardiac Implantable Devices and Noncardiac Mimics on Chest Radiographs.} \par
Cardiac implantable electronic devices (CIEDs), including pacemakers and defibrillators, are increasingly used to manage various cardiac conditions. This article reviews the radiographic appearance, typical components, and placement of CIEDs, including newer technologies like leadless pacemakers and MRI-conditional devices. The article also highlights the imaging findings of common complications such as lead dislodgement, fracture, and... (Omitted) \par
\small URL: https://pubmed.ncbi.nlm.nih.gov/40358448/ | Source: PubMed
\\
\midrule
\textbf{Textbook Summary 1} \par
Atelectasis is characterized by radiologic findings of increased lung density, crowding of pulmonary vessels, and displacement of fissures, hilum, and mediastinum, often with elevation of the hemidiaphragm. On MRI, obstructive atelectasis shows high signal intensity on T2WI, while nonobstructive atelectasis shows low signal intensity. The diagnosis of atelectasis considers various causes, including obstructive, nonobstructive, passive, adhesive, and cicatrizing types, each with distinct pathophysiology and imaging features.
\\
\midrule
\textbf{Textbook Summary 2} \par
Fracture is not a primary focus in the provided textbook pages, which mainly discuss lymphoma and malignant melanoma, but radiologic findings related to fractures are not mentioned. However, in the context of lymphoma, particularly enteropathy-associated T-cell lymphoma, complications such as bowel perforation can occur, which may be associated with a fracture of the bowel wall. Diagnostic considerations for lymphoma include imaging findings like circumferential wall thickening on CT and FDG uptake on PET.
\\
\midrule
\textbf{Assessment Question 1} \par
\textit{What is a characteristic radiologic finding of atelectasis?} \par
\begin{itemize}
    \item A. Decreased lung density
    \item B. Increased lung density
    \item C. Dilatation of pulmonary vessels
    \item D. Elevation of the diaphragm without crowding of pulmonary vessels
\end{itemize}
\textbf{Answer:} B \par
\textbf{Explanation:} Atelectasis is characterized by radiologic findings of increased lung density, crowding of pulmonary vessels, and displacement of fissures, hilum, and mediastinum.
\\
\midrule
\textbf{Assessment Question 2} \par
\textit{Which of the following statements about postoperative atelectasis is supported by the provided context?} \par
\begin{itemize}
    \item A. Atelectasis is a common cause of postoperative fever.
    \item B. Atelectasis is rarely seen in postoperative patients.
    \item C. The belief that atelectasis causes postoperative fever is a pervasive myth.
    \item D. Atelectasis always leads to significant respiratory complications.
\end{itemize}
\textbf{Answer:} C \par
\textbf{Explanation:} The context mentions that the belief that postoperative atelectasis causes fever is a myth, despite ample evidence to the contrary, as discussed in the abstract titled "Debunking a mythology: Atelectasis is not a cause of postoperative fever."
\\
\midrule
\textbf{Assessment Question 3} \par
\textit{What is the sensitivity of FRACTURE-MRI in detecting fractures, as mentioned in the provided context?} \par
\begin{itemize}
    \item A. 70%
    \item B. 80%
    \item C. 91.5%
    \item D. 95%
\end{itemize}
\textbf{Answer:} C \par
\textbf{Explanation:} The context states that fractures were detected by FRACTURE-MRI with a sensitivity of 91.5
\\
\midrule
\textbf{Assessment Question 4} \par
\textit{What is a potential advantage of using FRACTURE-MRI for diagnosing proximal tibial fractures, as mentioned in the context?} \par
\begin{itemize}
    \item A. Higher radiation exposure
    \item B. Reduced accuracy in fracture classification
    \item C. Reduced costs and radiation exposure
    \item D. Increased need for CT scans
\end{itemize}
\textbf{Answer:} C \par
\textbf{Explanation:} The context suggests that FRACTURE-MRI might be a helpful alternative to computed tomography in an acute trauma setting by reducing costs and radiation exposure in patients requiring a preoperative MRI anyway.
\\
\midrule
\textbf{Final Educational Material} \par
\par\textbf{Atelectasis}\par \par Atelectasis is characterized by increased lung density, crowding of pulmonary vessels, and displacement of fissures, hilum, and mediastinum, often with elevation of the hemidiaphragm. It's essential to differentiate atelectasis from consolidation, as their imaging characteristics and clinical implications differ. Atelectasis can be caused by various factors, including obstructive, nonobstructive, passive, adhesive, and cicatrizing types. Notably, atelectasis is not a cause of postoperative fever, contrary to a common myth. The use of artificial intelligence in triaging chest radiographs can help reduce the workload of radiologists while maintaining sensitivity for detecting atelectasis and other radiologic findings. \par \par\textbf{Fracture}\par \par Fracture diagnosis and classification can be accurately performed using 3D FRACTURE (fast field echo resembling a CT using restricted echo-spacing) MRI sequences, which offer a feasible alternative to CT scans in acute trauma settings. This can help reduce radiation exposure and costs, especially in patients requiring preoperative MRI. When evaluating fractures, particularly in the context of cardiac implantable devices, it's crucial to identify potential complications such as lead dislodgement, fracture, and perforation. Accurate identification of these devices and their positioning on chest radiographs is vital for early detection of complications and directing appropriate patient care.
\\
\end{longtable}
\end{feedbackbox}


\clearpage

\section{Default System Prompts for \ours}
\label{sec:appendix_prompts}

\subsection{Keyword Generation Prompt}
\begin{promptbox}[title=Keyword Generation Prompt]
\label{sec: keyword generation prompt}
\textbf{System Prompt:}
\begin{verbatim}
You are an expert medical language model. Given the full radiology report 
and the extracted Impression section, extract all specific disease names, 
diagnostic labels, and named pathological entities mentioned or implied in 
either section. Focus only on established or suspected diagnoses, such as 
named conditions.

Only include diagnoses that are positively identified or suspected in the 
report. Do not include any conditions that are explicitly ruled out, 
negated, or stated as absent.

Do not include general phrases, symptoms, or clinical findings that are not 
formal diagnoses.

Output your answer as a valid JSON object with the following format:

{ "keywords": ["diagnosis 1", "diagnosis 2", "diagnosis 3"] }

If no diagnoses are present, return:

{
"keywords": []
}
\end{verbatim}

\vspace{2mm}
\textbf{User Instruction Template:}
\begin{verbatim}
Final_report: {full_report_text}
Impression: {impression_text}
\end{verbatim}
\end{promptbox}

\subsection{Textbook Summary Prompt}
\begin{promptbox}[title=Textbook Summary Prompt]
\label{sec: textbook summary prompt}
\textbf{System Prompt:}
\begin{verbatim}
You are a concise and accurate radiology assistant, skilled in summarizing 
medical texts.
\end{verbatim}

\vspace{2mm}
\textbf{User Instruction Template:}
\begin{verbatim}
Please summarize the following textbook pages focusing on the keyword '{keyword}'. 
The summary should highlight key radiologic findings and diagnostic 
considerations. Be concise, using 2-3 sentences and your own words. 
Output only the summary text itself, with no additional conversational 
text or headers.

Textbook Pages Content:
{pages_block_text}
\end{verbatim}
\end{promptbox}

\subsection{MCQ Generation Prompt}
\begin{promptbox}[title=Multiple Choice Q\&A Generation Prompt]
\label{sec: mcq prompt}
\textbf{System Prompt:}
\begin{verbatim}
You are a specialized AI assistant for creating multiple-choice questions (MCQs) 
for radiology education. You must focus *exclusively* on the provided 
**Primary Diagnostic Keywords**.
\end{verbatim}

\vspace{2mm}
\textbf{User Instruction Template:}
\begin{verbatim}
### Primary Diagnostic Keywords to Focus On:
- {keywords_list_str}

### Full Context (for reference)
{mcq_input_context}

### Your Task
Based *only* on the provided context, generate 2 multiple-choice questions 
**for each Primary Diagnostic Keyword listed above**. Do not generate questions 
for any other terms or topics mentioned in the context. Each question must 
test understanding of the information related to the primary keywords.

Follow this format exactly:

### Multiple Choice Questions

#### {{Diagnosis Keyword 1}}

Q1. {{Question stem}}
A. {{Option A}}
B. {{Option B}}
C. {{Option C}}
D. {{Option D}}
Answer: {{Correct Option Letter}}
Explanation: {{Brief explanation based on the provided context.}}
\end{verbatim}
\end{promptbox}

\subsection{Educational Material Generation Prompt}
\begin{promptbox}[title=Educational Material Prompt]
\label{sec: final feedback prompt}
\textbf{System Prompt:}
\begin{verbatim}
You are an expert radiology AI assistant. Your task is to synthesize the 
provided information into concise, educational feedback focused *only* on the 
primary diagnostic keywords provided. Do not explain or elaborate on other 
terms from the original report unless they are directly relevant to the 
primary keywords.
\end{verbatim}

\vspace{2mm}
\textbf{User Instruction Template:}
\begin{verbatim}
### Primary Diagnostic Keywords
- {keywords_list_str}

### Original Reviewer Report (for context only)
{original_reviewer_report}

### Supporting Educational Material
{user_block_for_final_stages}

### Your Task
Based on all the information above, provide a concise, synthesized feedback. 
Structure your response with a section for each **Primary Diagnostic Keyword**. 
Focus only on clinical teaching points and imaging pearls related to these 
primary keywords.
\end{verbatim}
\end{promptbox}

\clearpage

\section{MedTutor System UI}
\label{sec: medtutor ui}

\begin{figure}[h!]
    \centering

    \begin{subfigure}[b]{0.9\textwidth}
        \centering
        \includegraphics[width=\linewidth]{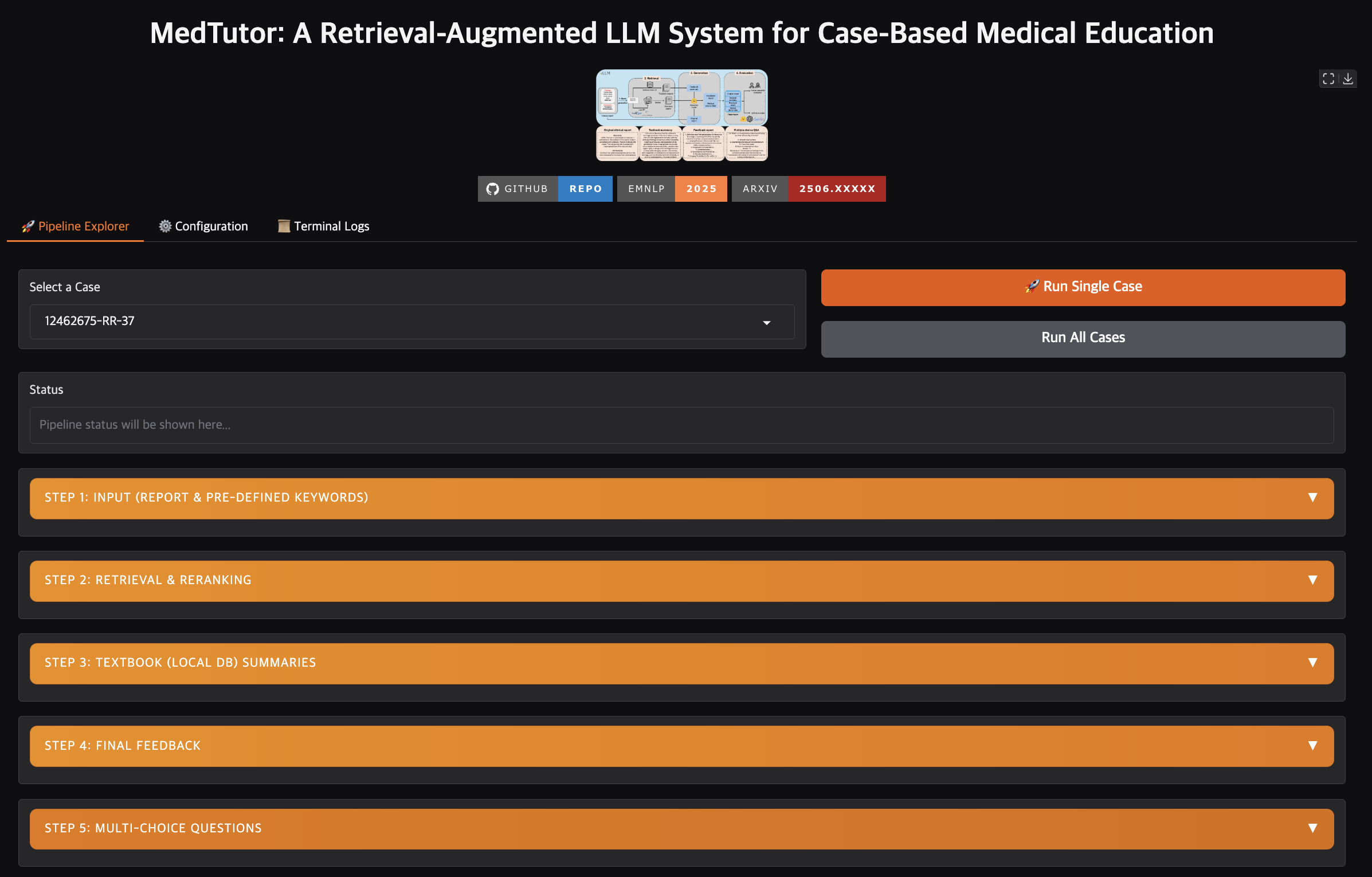}
        \caption{Main user interface of MedTutor.}
        \label{subfig:medtutor-main}
    \end{subfigure}

    \vspace{1.5cm}

    \begin{subfigure}[b]{0.9\textwidth}
        \centering
        \includegraphics[width=\linewidth]{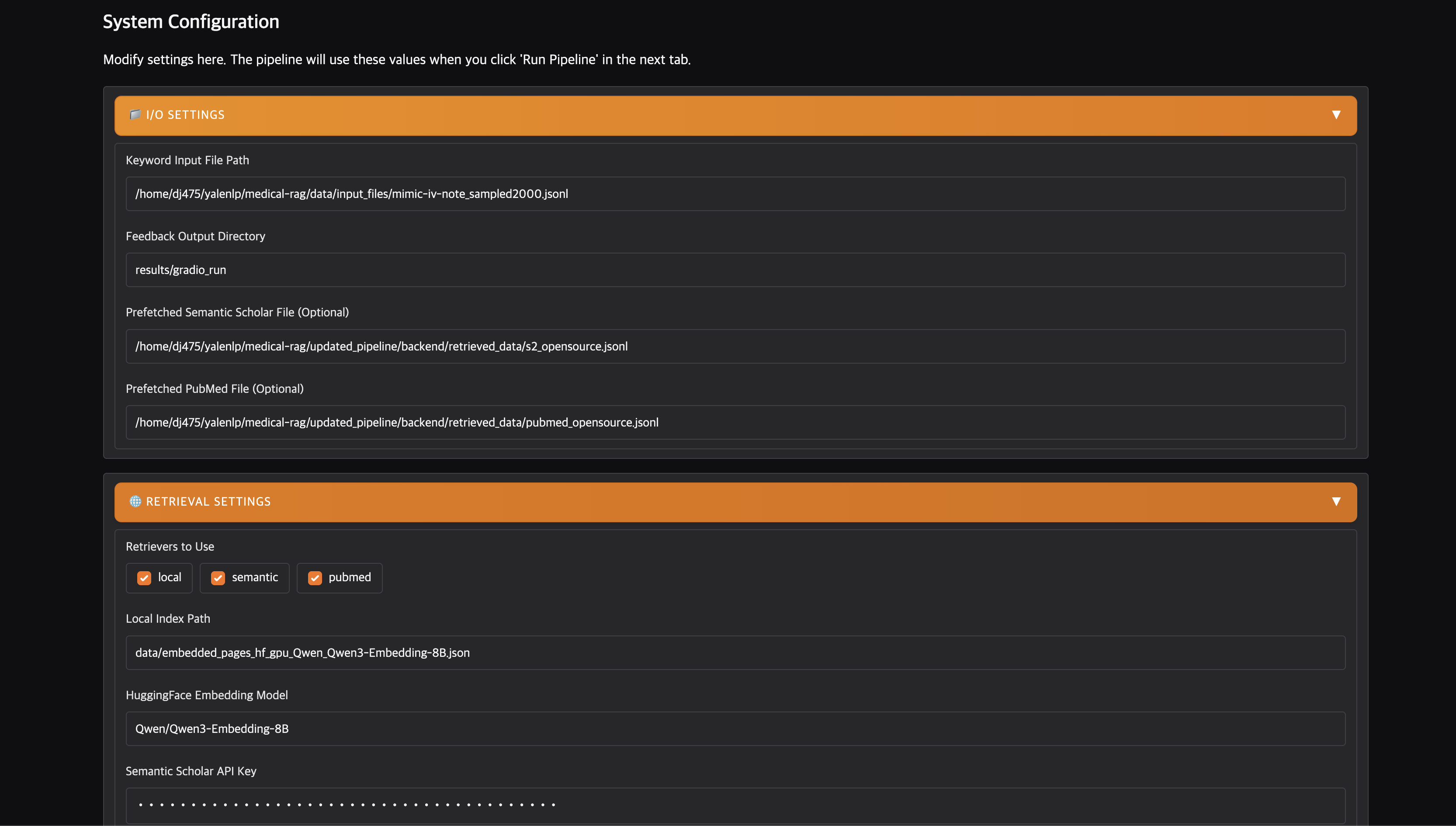}
        \caption{Configuration settings for model selection and system prompts.}
        \label{subfig:medtutor-config1}
    \end{subfigure}

    \caption{The MedTutor UI (Part 1 of 2): Main dashboard and initial configuration settings.}
    \label{fig:medtutor-ui-part1}
\end{figure}

\clearpage

\begin{figure}[p]
    \centering

    \begin{subfigure}[b]{0.9\textwidth}
        \centering
        \includegraphics[width=\linewidth]{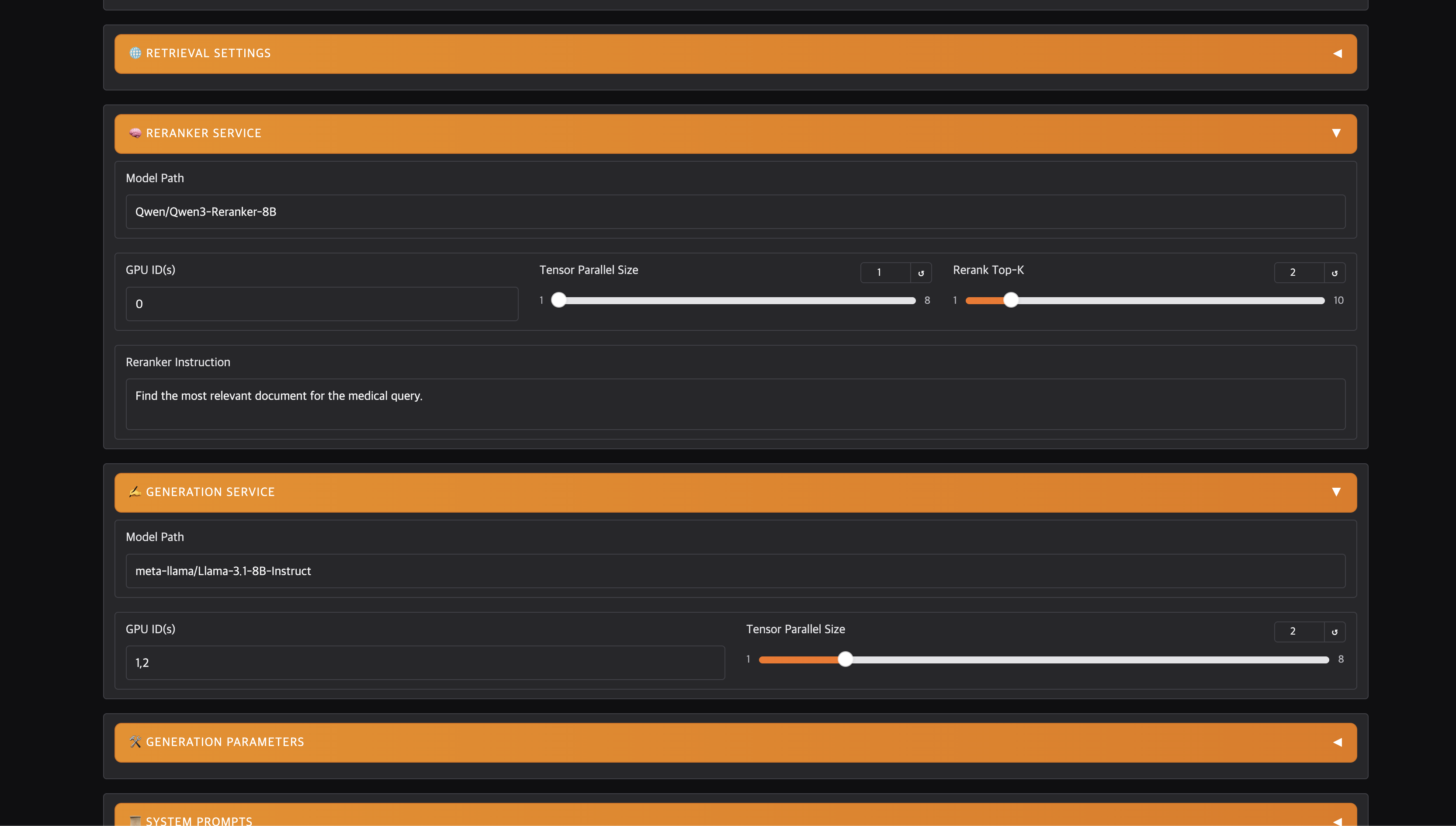}
        \caption{Further configuration for data sources and retrieval.}
        \label{subfig:medtutor-config2}
    \end{subfigure}

    \vspace{1.5cm}

    \begin{subfigure}[b]{0.9\textwidth}
        \centering
        \includegraphics[width=\linewidth]{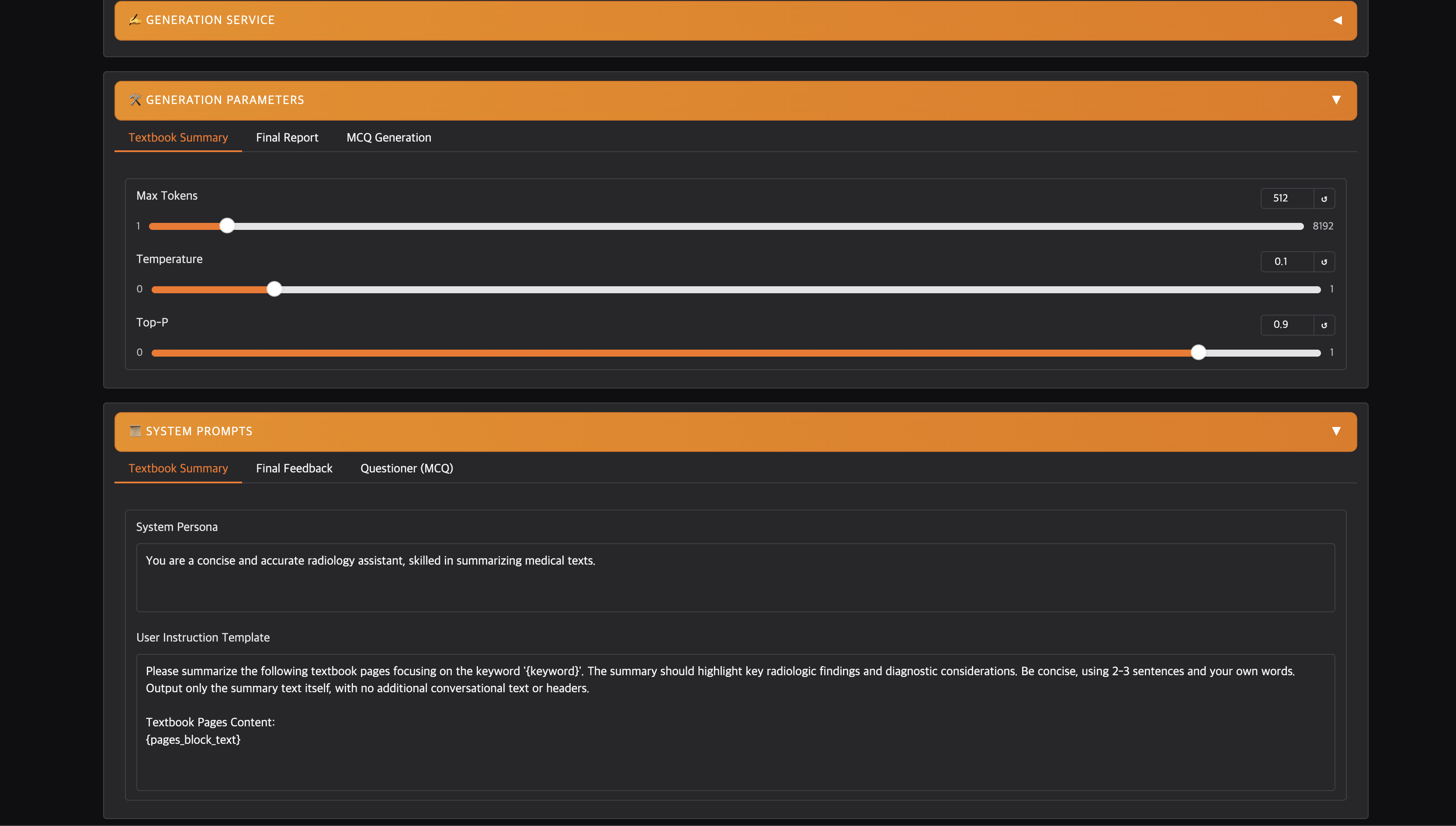}
        \caption{Finalizing configuration and execution options.}
        \label{subfig:medtutor-config3}
    \end{subfigure}

    \caption{The MedTutor UI (Part 2 of 2): Additional configuration panels for data processing and task execution.}
    \label{fig:medtutor-ui-part2}
\end{figure}

\clearpage

\section{Human Annotation Guideline}
\label{sec:annotation guideline}

\begin{infobox}[title=I. Evaluation of Information Quality per Keyword]
For each diagnostic keyword identified from the original report, we evaluate the following components:

\subsection{Retrieved \& Reranked Academic Papers}
\begin{itemize}[leftmargin=*, itemsep=0pt]
    \item \textbf{Relevance to Keyword \& Original Report:} How directly related is each paper or retrieved snippet to the given keyword and the context of the original radiology report?
\end{itemize}

\subsection{Generated Textbook Summary}
\begin{itemize}[leftmargin=*, itemsep=0pt]
    \item \textbf{Accuracy \& Factuality:} Is the summary an accurate and factual representation of information related to the keyword (compared to general radiology knowledge or, if available, the source textbook)?
    \item \textbf{Helpfulness \& Relevance:} Is the summary helpful and related to the case report provided as input? 
    \item \textbf{Coverage of Key Information:} Does the summary include the most critical information (e.g., key imaging findings, diagnostic criteria) related to the keyword?
\end{itemize}

\subsection{Example Multiple Choice Questions}
\begin{itemize}[leftmargin=*, itemsep=0pt]
    \item \textbf{Relevance \& Correctness:} Are the questions relevant to the keyword? Is the answer provided and rationale correct? Are the answer choices relevant?
\end{itemize}
\end{infobox}

\vspace{4mm} 

\begin{infobox}[title=II. Evaluation of the "Educational material" Paragraph (per Keyword)]
\begin{itemize}[leftmargin=*, itemsep=0pt]
    \item \textbf{Clinical \& Educational Utility:} How clinically relevant, accurate, and educationally valuable is this educational material paragraph for a radiology trainee in understanding the keyword within the context of the original report? (This encompasses quality, clinical insight, contextual appropriateness, and trustworthiness.)
\end{itemize}
\end{infobox}

\vspace{4mm}

\begin{infobox}[title=III. Evaluation of Overall Educational Material Structure \& Quality]
For the entire generated report:
\begin{itemize}[leftmargin=*, itemsep=0pt]
    \item \textbf{Appropriateness of Keywords:} Are the keywords (used to structure the feedback) appropriate and comprehensive for the given original radiology report? Specifically, is the keyword general enough that it can be searched in a textbook or Radiopaedia (e.g., “rib fracture”, not “anterior 4th rib fracture”), and related to a pathology worth learning more about (e.g., “cholangiocarcinoma”, not “mass”)?
\end{itemize}
\end{infobox}

\clearpage

\section{Human Annotator System UI}
\label{sec:annotator system ui}

\begin{figure}[h!]
    \centering
  
    \begin{subfigure}[b]{0.9\textwidth} 
        \centering
        \includegraphics[width=\linewidth]{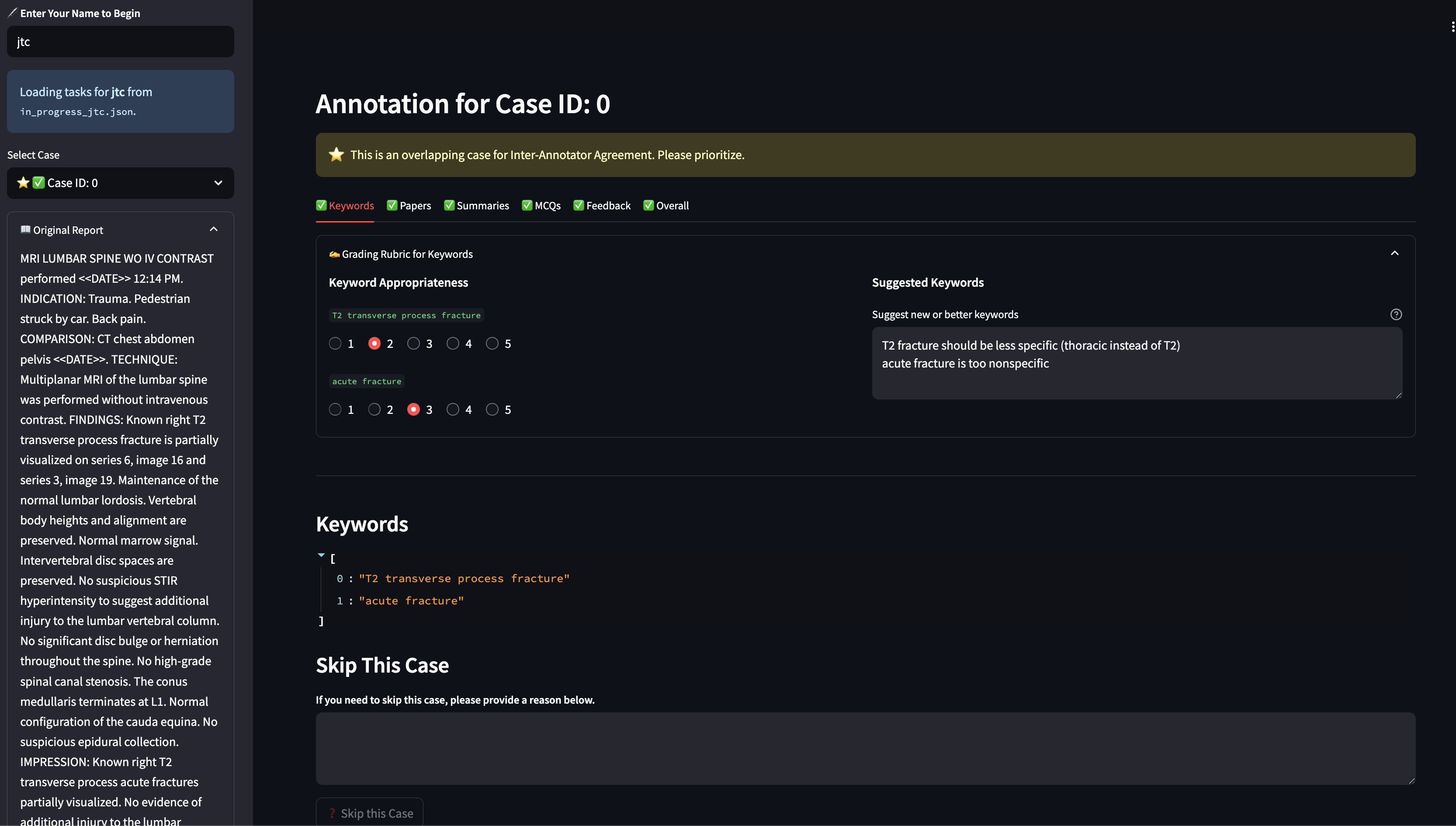}
        \caption{Keyword Evaluation Page}
        \label{subfig:keyword-eval}
    \end{subfigure}

    \vspace{1.5cm} 

    \begin{subfigure}[b]{0.9\textwidth}
        \centering
        \includegraphics[width=\linewidth]{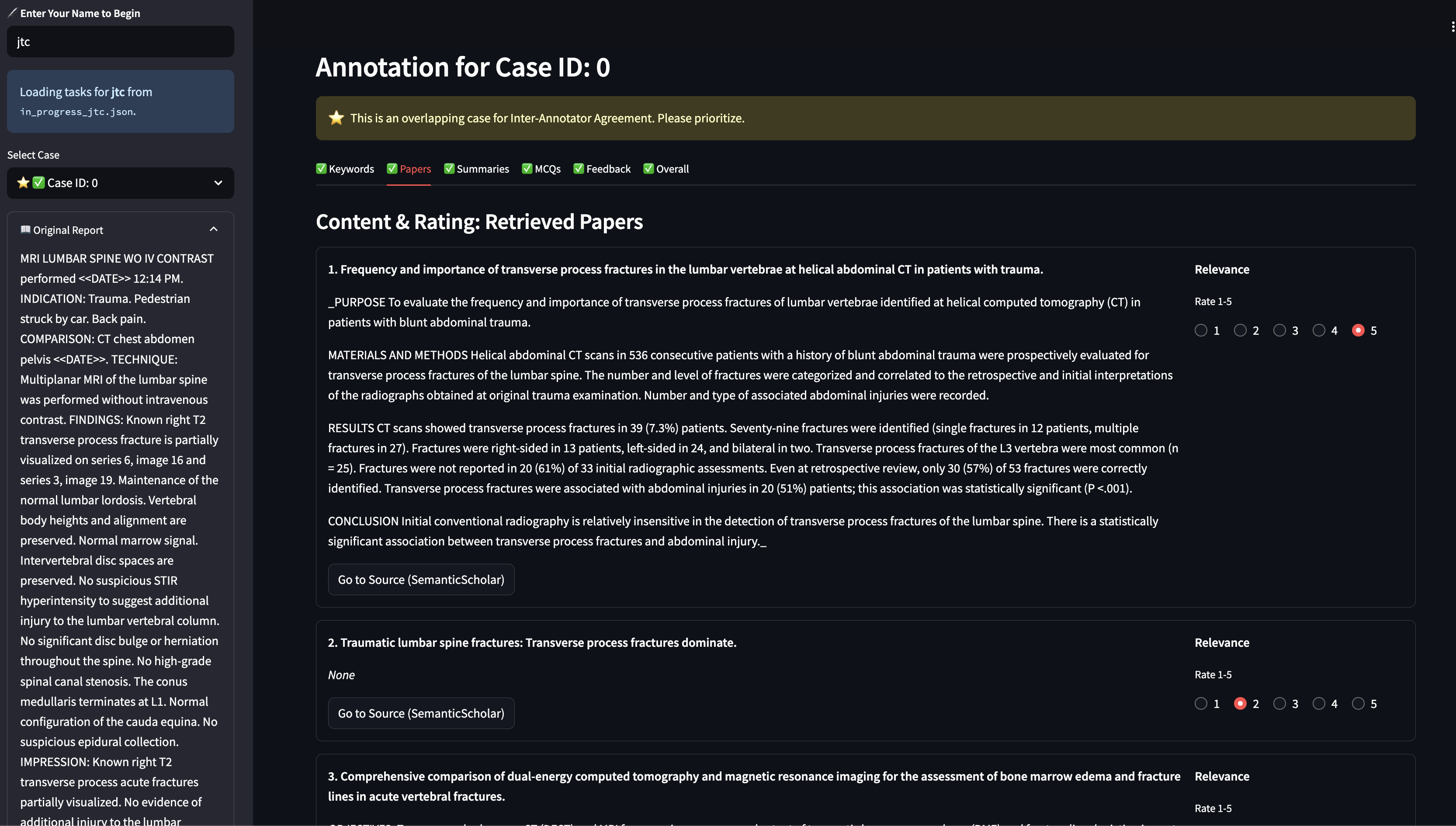}
        \caption{Paper Evaluation Page}
        \label{subfig:paper-eval}
    \end{subfigure}

    \caption{Annotation system UI (Part 1 of 3): Interfaces for evaluating keywords and retrieved papers.}
    \label{fig:annotation-ui-part1}
\end{figure}

\clearpage

\begin{figure}[p]
    \centering

    \begin{subfigure}[b]{0.9\textwidth}
        \centering
        \includegraphics[width=\linewidth]{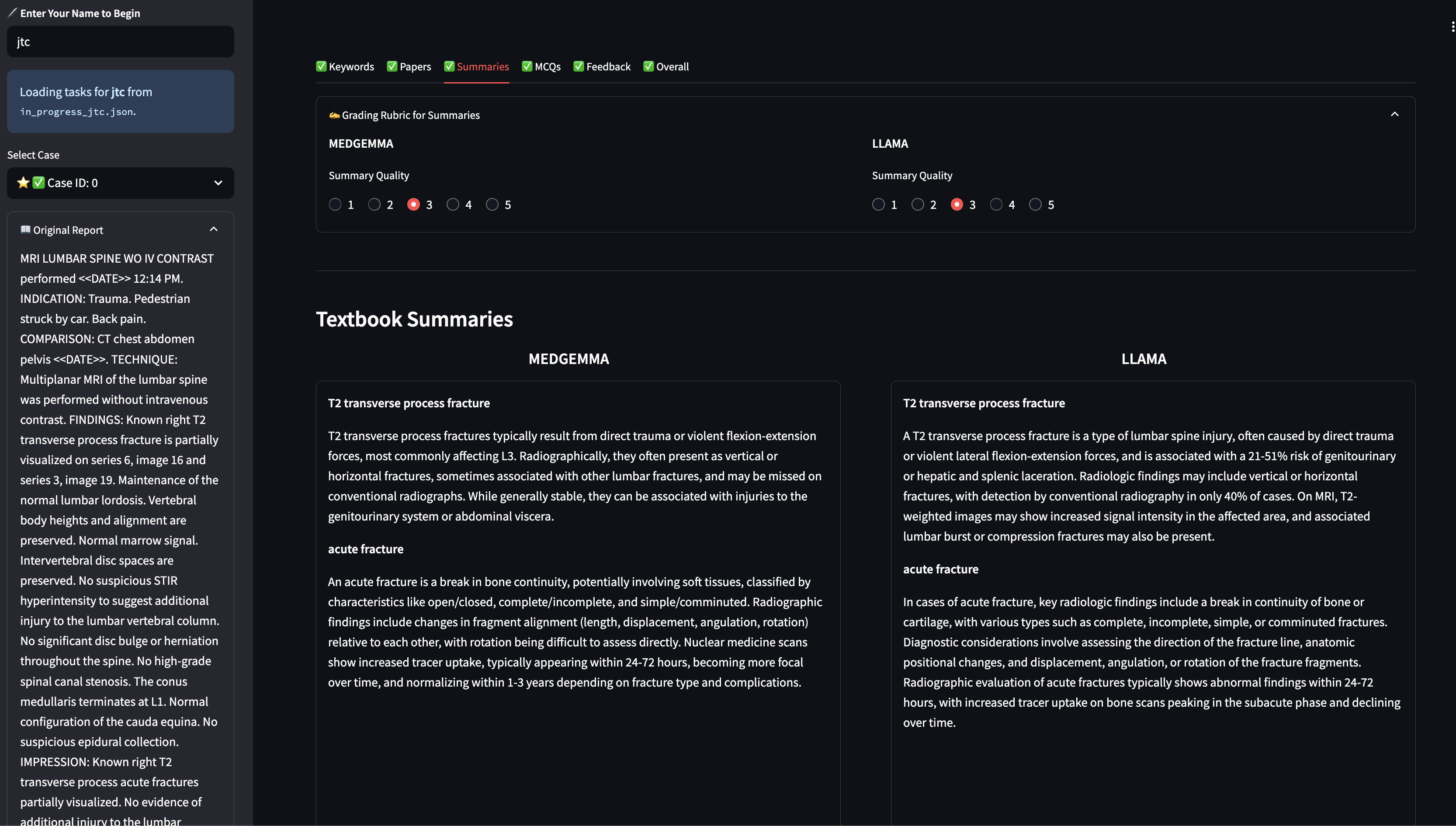}
        \caption{Textbook Summary Evaluation Page}
        \label{subfig:textbook-eval}
    \end{subfigure}

    \vspace{1.5cm}

    \begin{subfigure}[b]{0.9\textwidth}
        \centering
        \includegraphics[width=\linewidth]{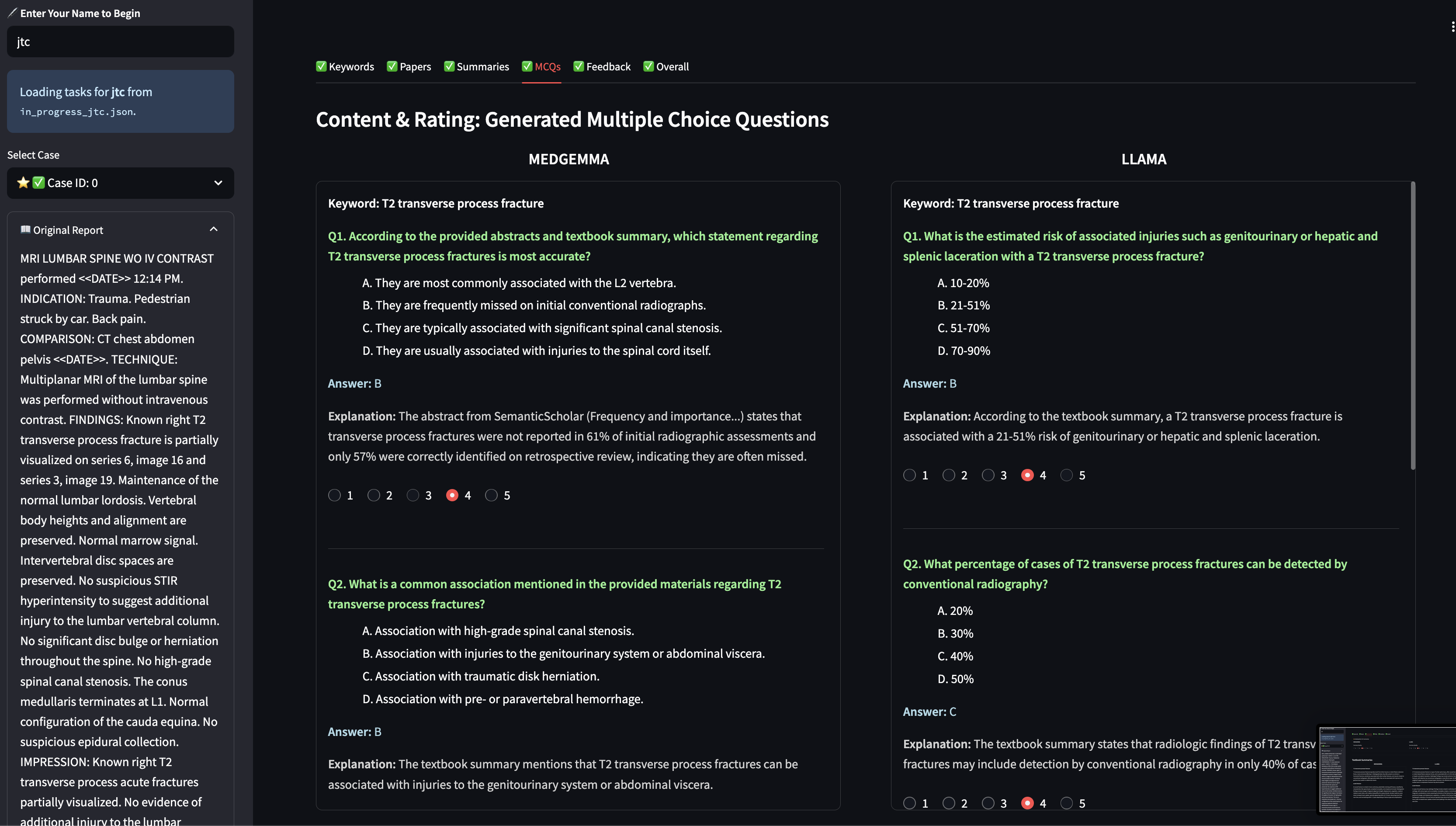}
        \caption{MCQ Evaluation Page}
        \label{subfig:mcq-eval}
    \end{subfigure}

    \caption{Annotation system UI (Part 2 of 3): Interfaces for evaluating textbook summaries and multiple-choice questions.}
    \label{fig:annotation-ui-part2}
\end{figure}

\clearpage

\begin{figure}[p]
    \centering

    \begin{subfigure}[b]{0.9\textwidth}
        \centering
        \includegraphics[width=\linewidth]{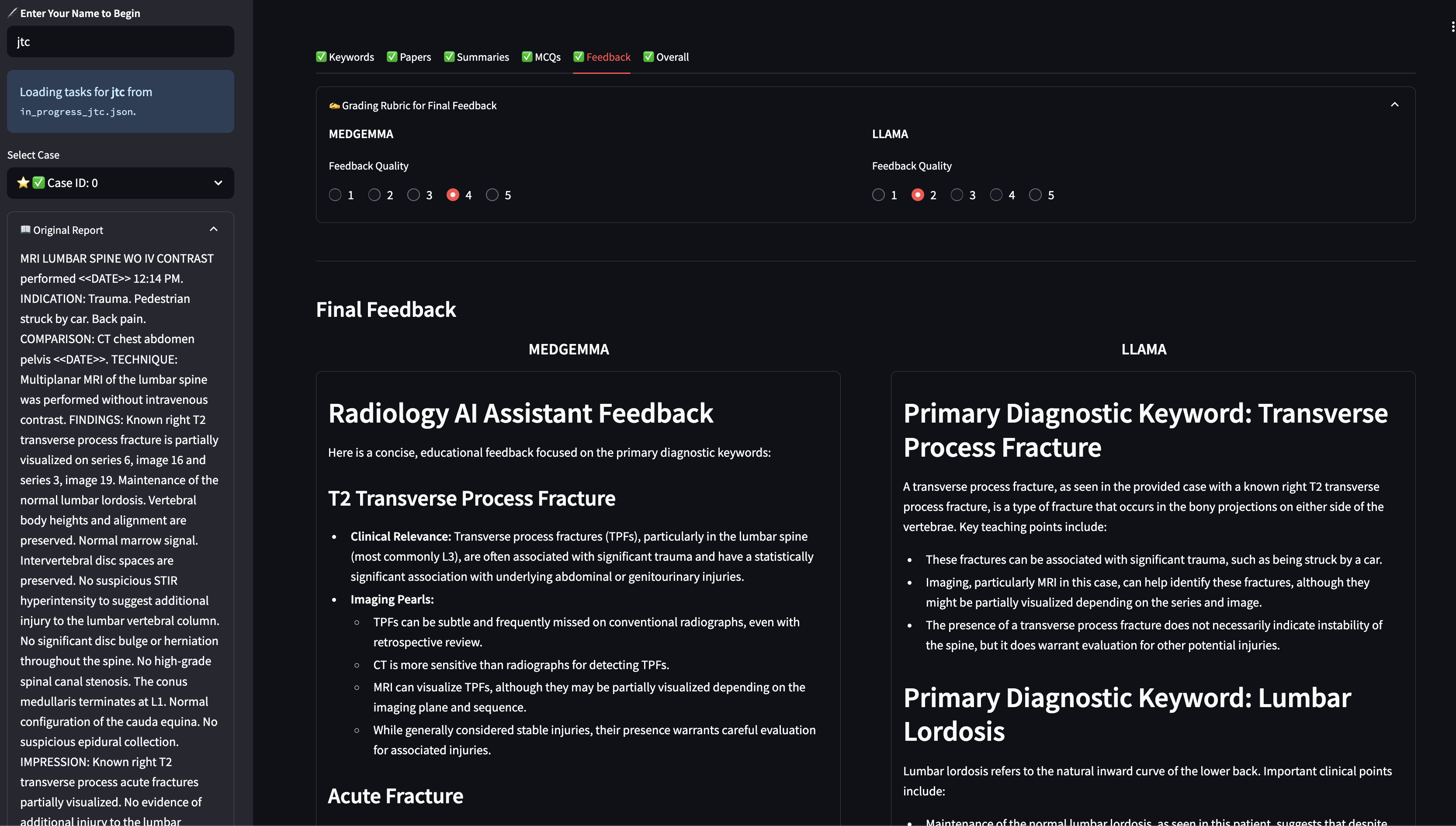}
        \caption{Educational Material Evaluation Page}
        \label{subfig:feedback-eval}
    \end{subfigure}

    \vspace{1.5cm}

    \begin{subfigure}[b]{0.9\textwidth}
        \centering
        \includegraphics[width=\linewidth]{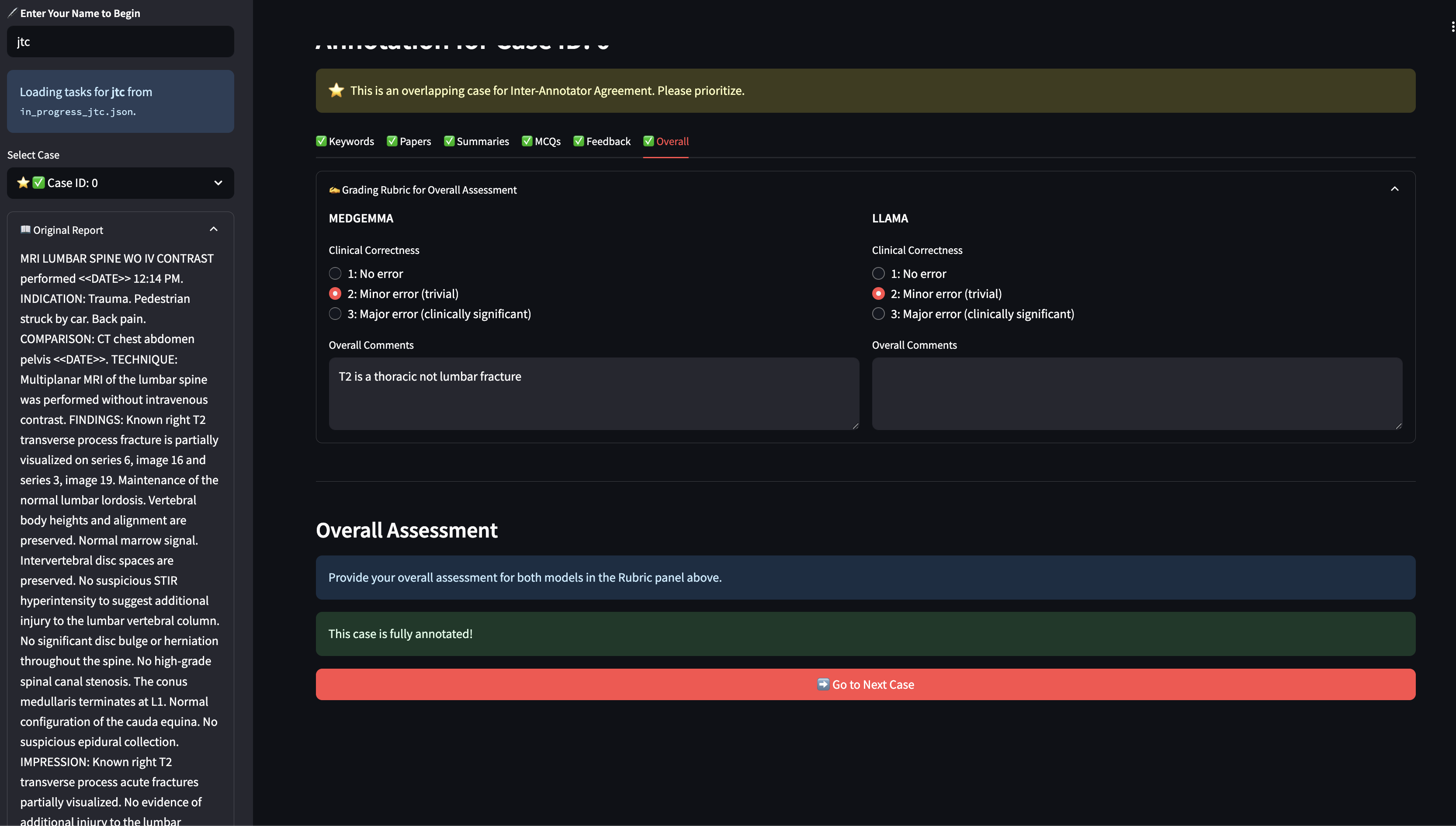}
        \caption{Overall Quality Evaluation Page}
        \label{subfig:overall-eval}
    \end{subfigure}

    \caption{Annotation system UI (Part 3 of 3): Interfaces for evaluating the final synthesized educational material and overall quality.}
    \label{fig:annotation-ui-part3}
\end{figure}

\end{document}